\documentclass{article}
\usepackage{authblk}
\usepackage[utf8]{inputenc}
\usepackage{main}
\usepackage{microtype}
\usepackage{subcaption}
\usepackage{graphicx}
\usepackage{times}
\usepackage{latexsym}
\usepackage{amsmath}
\usepackage{float}
\usepackage{footnote}
\usepackage{enumitem}
\usepackage{bm}
\usepackage{arydshln}
\usepackage{booktabs}
\usepackage{array}     
\usepackage{multicol}
\usepackage{multirow}
\usepackage{color}
\usepackage{xcolor}     
\usepackage{colortbl}
\usepackage{bbding}
\usepackage{makecell}
\usepackage{mathtools}
\usepackage{imakeidx}
\usepackage{longtable}
\usepackage{wrapfig}
\usepackage{algorithmic}
\usepackage{rotating}
\makeindex
\usepackage{arydshln}
\usepackage{lipsum}
\usepackage{natbib}
\usepackage[edges]{forest}
\usepackage[normalem]{ulem}
\definecolor{mydarkblue}{rgb}{0,0.08,0.45}
\usepackage[colorlinks=true,linkcolor=mydarkblue,citecolor=mydarkblue,filecolor=mydarkblue,urlcolor=mydarkblue]{hyperref}
\usepackage{CJKutf8}
\usepackage{awesomebox} 
\usepackage{bbding}
\usepackage[most]{tcolorbox}
\usepackage{booktabs}
\usepackage{geometry}
\definecolor{wkblue}{rgb}{0.2, 0.3, 0.6}
\definecolor{meta-color}{rgb}{0.5, 0.5, 0.5}
\usepackage{amsmath}
\usepackage{enumitem}
\usepackage{lscape} 
\usepackage{booktabs}
\usepackage{algorithm}   
\usepackage{setspace}

\usepackage{algorithmic}
\usepackage{tabularx,booktabs}
\usepackage{makecell}

\usepackage{amssymb}
\usepackage{amsfonts}

\usepackage{booktabs} 
\usepackage{geometry} 

\usepackage{multirow}

\usepackage[tikz]{bclogo}
\usepackage[framemethod=tikz]{mdframed}
\definecolor{bgblue}{RGB}{245,243,253}
\definecolor{ttblue}{RGB}{91,194,224}

\usepackage{pgfplots}
\usepackage{pgfplotstable}

\usepackage{wrapfig}
\usepackage{graphicx}

\usepackage{booktabs, multirow} 
\usepackage{soul}
\usepackage{xcolor,colortbl} 
\usepackage{changepage,threeparttable} 

\usepackage{tcolorbox}

\usepackage[most]{tcolorbox}
\tcbuselibrary{skins,breakable,listings}
\usepackage{fvextra} 
\usepackage{amsmath, amssymb, amsthm}

\DefineVerbatimEnvironment{TextBlock}{Verbatim}{
  breaklines=true,
  breakanywhere=true,     
  breaksymbolleft={},     
  obeytabs=true,
  commandchars=\\\{\}     
}

\newtcolorbox{CompareBox}[2][]{%
  enhanced, breakable,
  colback=#2!5!white,
  colframe=#2!50!black,
  boxrule=1pt,
  arc=6pt,
  top=6pt, bottom=6pt, left=6pt, right=6pt,
  title=#1,
  fonttitle=\bfseries,
}

\newcommand{\added}[1]{\textcolor{purple!70!black}{#1}}

\mdfdefinestyle{mystyle}{%
  rightline=true,
  innerleftmargin=10,
  innerrightmargin=10,
  outerlinewidth=3pt,
  topline=false,
  rightline=true,
  bottomline=false,
  skipabove=\topsep,
  skipbelow=\topsep
}

\newtcolorbox{keybox}{
    colback=blue!5!white,
    colframe=blue!75!black,
    fonttitle=\bfseries,
    title=Takeaways
}

\newtcolorbox{myboxi}[1][]{
  breakable,
  title=#1,
  colback=red!5,
  colbacktitle=red!5,
  coltitle=black,
  fonttitle=\bfseries,
  bottomrule=0pt,
  toprule=0pt,
  leftrule=2pt,
  rightrule=2pt,
  titlerule=0pt,
  arc=0pt,
  outer arc=0pt,
  colframe=red,
}

\newtcolorbox{myboxnote}[1][]{
  breakable,
  title=#1,
  colback=orange!0,
  colbacktitle=orange!0,
  coltitle=black,
  fonttitle=\bfseries,
  bottomrule=0pt,
  toprule=0pt,
  leftrule=2pt,
  rightrule=2pt,
  titlerule=0pt,
  arc=0pt,
  outer arc=0pt,
  colframe=orange,
}

\newtcolorbox{myboxii}[1][]{
  breakable,
  freelance,
  title=#1,
  colback=white,
  colbacktitle=white,
  coltitle=black,
  fonttitle=\bfseries,
  bottomrule=0pt,
  boxrule=0pt,
  colframe=white,
  overlay unbroken and first={
  \draw[red!75!black,line width=3pt]
    ([xshift=5pt]frame.north west) -- 
    (frame.north west) -- 
    (frame.south west);
  \draw[red!75!black,line width=3pt]
    ([xshift=-5pt]frame.north east) -- 
    (frame.north east) -- 
    (frame.south east);
  },
  overlay unbroken app={
  \draw[red!75!black,line width=3pt,line cap=rect]
    (frame.south west) -- 
    ([xshift=5pt]frame.south west);
  \draw[red!75!black,line width=3pt,line cap=rect]
    (frame.south east) -- 
    ([xshift=-5pt]frame.south east);
  },
  overlay middle and last={
  \draw[red!75!black,line width=3pt]
    (frame.north west) -- 
    (frame.south west);
  \draw[red!75!black,line width=3pt]
    (frame.north east) -- 
    (frame.south east);
  },
  overlay last app={
  \draw[red!75!black,line width=3pt,line cap=rect]
    (frame.south west) --
    ([xshift=5pt]frame.south west);
  \draw[red!75!black,line width=3pt,line cap=rect]
    (frame.south east) --
    ([xshift=-5pt]frame.south east);
  },
}

\usepackage{fontawesome5}
\usepackage{fancyhdr} 
\usepackage{blindtext} 
\usepackage{makecell}

\DeclareCaptionFont{black}{\color{black}}

\definecolor{myblue}{rgb}{0.9, 0.1, 0.94}
\definecolor{mygreen}{rgb}{0.64, 0.56, 0.88}
\definecolor{myyellow}{rgb}{0.68, 0.6, 0.1}
\definecolor{fancygreen}{rgb}{0.33, 0.68, 0.20}
\definecolor{salmon}{rgb}{0.94, 0.52, 0.49}
\definecolor{tablegreen}{rgb}{0.82, 0.94, 0.75}
\definecolor{tableblue}{rgb}{0.81, 0.90, 0.94}
\definecolor{tablered}{rgb}{0.97, 0.85, 0.85}
\definecolor{tableorange}{rgb}{0.96, 0.85, 0.81}

\newenvironment{itemize*}%
 {\leftmargini=10pt\begin{itemize}%
  \setlength{\itemsep}{0pt}%
  \setlength{\parskip}{0pt}%
  }%
 {\end{itemize}}
\newenvironment{enumerate*}%
 {\begin{enumerate}%
  \setlength{\itemsep}{0pt}%
  \setlength{\parskip}{0pt}}%
 {\end{enumerate}}

\usepackage{xcolor}
\usepackage{listings}  

\newcommand\JSONnumbervaluestyle{\color{blue}}
\newcommand\JSONstringvaluestyle{\color{red}}

\newif\ifcolonfoundonthisline

\makeatletter

\lstdefinestyle{json}
{
  showstringspaces    = false,
  keywords            = {false,true},
  alsoletter          = 0123456789.,
  morestring          = [s]{"}{"},
  stringstyle         = \ifcolonfoundonthisline\JSONstringvaluestyle\fi,
  MoreSelectCharTable =%
    \lst@DefSaveDef{`:}\colon@json{\processColon@json},
  basicstyle          = \ttfamily,
  keywordstyle        = \ttfamily\bfseries,
}

\newcommand\processColon@json{%
  \colon@json%
  \ifnum\lst@mode=\lst@Pmode%
    \global\colonfoundonthislinetrue%
  \fi
}

\lst@AddToHook{Output}{%
  \ifcolonfoundonthisline%
    \ifnum\lst@mode=\lst@Pmode%
      \def\lst@thestyle{\JSONnumbervaluestyle}%
    \fi
  \fi
  \lsthk@DetectKeywords%
}

\lst@AddToHook{EOL}%
  {\global\colonfoundonthislinefalse}

\makeatother

\usepackage{etoolbox}
\usepackage{natbib}
\usepackage{url}
\newcounter{bibcount}
\makeatletter
\patchcmd{\@lbibitem}{\item[}{\item[\hfil\stepcounter{bibcount}{[\thebibcount]}}{}{}
\renewcommand\NAT@bibsetup%
  [1]{\setlength{\leftmargin}{\bibhang}\setlength{\itemindent}{-\parindent}%
      \setlength{\itemsep}{\bibsep}\setlength{\parsep}{\z@}}
\makeatother

\definecolor{mybrown}{RGB}{128,64,0}

\definecolor{titlecolor}{HTML}{4c9cff}
\definecolor{coolblue3}{rgb}{0.91, 0.94, 0.98}

\begin{document}

\title{Data Darwinism -- Part I\\Unlocking the Value of Scientific Data for Pre-training}

\author[1,2,4]{Yiwei Qin\textsuperscript{*}}
\author[1,3,4]{Zhen Huang\textsuperscript{*}}
\author[1,3,4]{Tiantian Mi\textsuperscript{*}}
\author[1,2,4]{Weiye Si}
\author[1,2,4]{Chenyang Zhou}
\author[1]{\authorcr Qipeng Guo}
\author[1]{Siyuan Feng}
\author[1,2,4]{Pengfei Liu\textsuperscript{†}}
\affil{SII \quad \textsuperscript{2}SJTU \quad \textsuperscript{3}FDU \quad \textsuperscript{4}GAIR}

\maketitle
\pagestyle{fancy}
\thispagestyle{fancy}
\fancyhead{}
\lhead{
  \raisebox{-0.3cm}{\includegraphics[height=0.95cm]{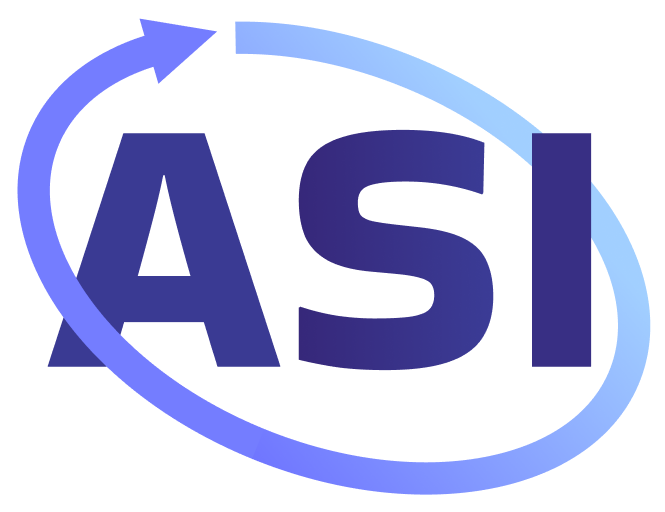}}
}
\rhead{%
  \raisebox{-0.2cm}{\includegraphics[height=0.7cm]{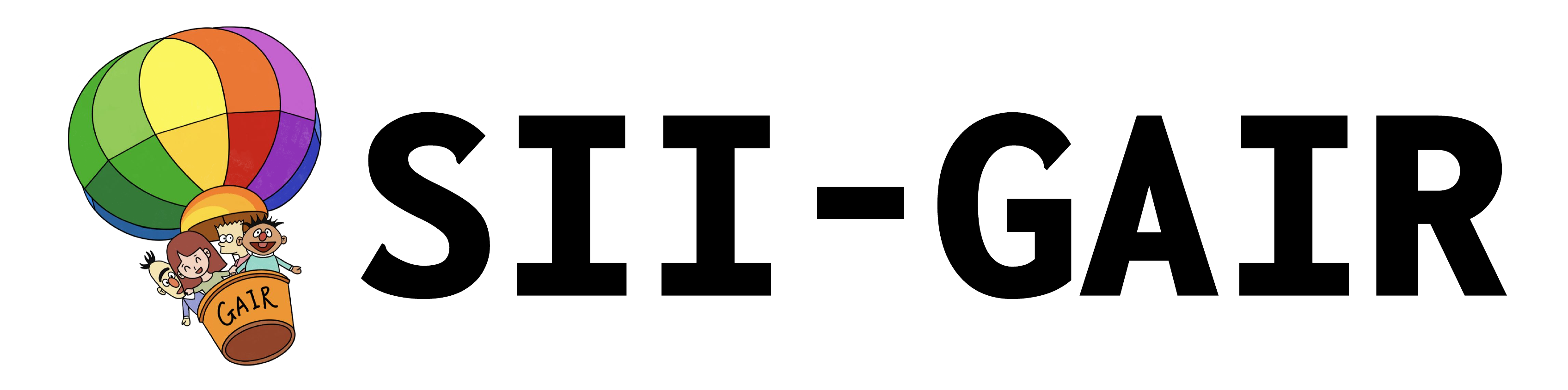}}%
}
\renewcommand{\headrulewidth}{0pt}
\setlength{\headsep}{2mm} 


\renewcommand{\thefootnote}{}
\footnotetext{* Equal contribution.}
\footnotetext{† Corresponding author.}
\vspace{-20pt}


{\centering
\quad \href{https://github.com/GAIR-NLP/Data-Darwinism}{\textcolor{black}\faGithub\ Data-Darwinism}
\quad \href{https://huggingface.co/GAIR/daVinci-origin-3B}{\raisebox{-.15em}{\includegraphics[height=1em]{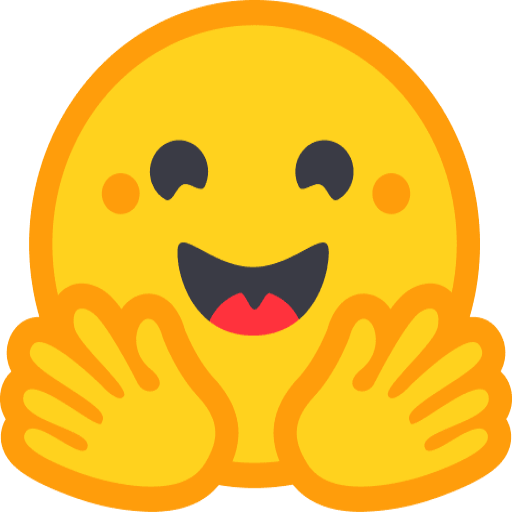}}\ daVinci-origin-3B}
\quad \href{https://huggingface.co/GAIR/daVinci-origin-7B}{\raisebox{-.15em}{\includegraphics[height=1em]{figs/huggingface-icon.png}}\ daVinci-origin-7B}
\\
\quad \href{https://huggingface.co/datasets/GAIR/Darwin-Science}{{\textcolor{violet}\faDatabase}\ Darwin-Science}
\quad \href{https://huggingface.co/datasets/GAIR/Darwin-Science-Eval}{{\textcolor{violet}\faDatabase}\ Darwin-Science-Eval}
\par}

\vspace{10pt}

\begin{abstract}
The quality of training data fundamentally determines foundation model performance, yet the field lacks systematic frameworks for data processing. We introduce \textbf{Data Darwinism}, a ten-level hierarchical taxonomy (L0--L9) organizing data transformations from selection to generation, preservation to transformation, and human-centric to machine-driven processing. This framework conceptualizes data as \textbf{co-evolving} with models: advanced models enable sophisticated processing, which produces superior training data for next-generation systems.

We validate this framework on scientific literature---a conceptually dense domain underutilized in open-source pre-training. We construct \texttt{Darwin-Science}, a 900B-token corpus implementing hierarchy levels L0--L5. Our key finding: raw scientific data suffers a severe \textbf{learnability gap}, providing negligible gains despite information density. We bridge this through L4 (Generative Refinement)---removing noise and repairing fragmentation---and L5 (Cognitive Completion)---expanding implicit reasoning, explicating terminology, and adding pedagogical bridges via frontier LLMs.
We establish rigorous controlled experiments with \texttt{Darwin-Science-Eval} (150K expert-level questions) and \textit{daVinci-origin-3B/7B}---which we pre-train \textbf{entirely from scratch} on 5.37T tokens deliberately excluding scientific content, a substantial undertaking enabling contamination-free baselines and unambiguous attribution of gains to data processing rather than checkpoint artifacts. Through 600B continued pre-training tokens, \texttt{Darwin-Science} outperforms baselines by \textbf{+2.12 (3B) and +2.95 (7B) points} on 20+ benchmarks, amplifying to \textbf{+5.60 and +8.40 points} on domain-aligned evaluation. Hierarchy progression from L0 to L5 yields \textbf{+1.36 total gain}, with L5 contributing +0.98, confirming systematic ascension unlocks latent value.
We release \texttt{Darwin-Science} and \textit{daVinci-origin-3B/7B} models to enable principled, co-evolutionary data-model development.


\end{abstract}

\begin{figure}[H]
    \centering
    \includegraphics[width=0.95\linewidth]{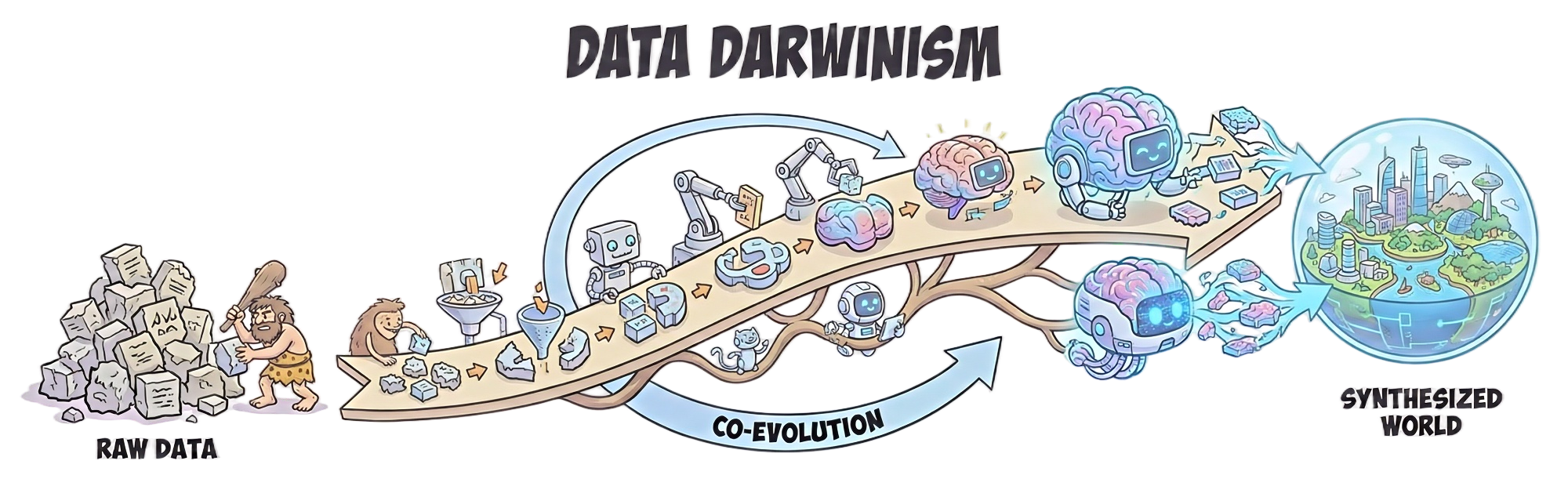}
    \caption{The Data Darwinism Pipeline. An evolutionary trajectory of data processing, illustrating the transition from raw data acquisition through model-driven refinement to the final stage of synthesized world generation.}
    \label{fig:data_darwinism_overview}
\end{figure}


\newpage

\pagestyle{fancy}
\lhead{\rightmark}
\renewcommand{\headrulewidth}{0.7pt}
\setlength{\headsep}{5mm}

\clearpage

\newpage

\renewcommand{\thefootnote}{\arabic{footnote}}
\setcounter{footnote}{0}  

\newpage
\onecolumn
\begin{spacing}{1.1}
\tableofcontents
\end{spacing}
\newpage

\section{Introduction}

The performance of foundation models is fundamentally determined by their training data~\citep{kaplan2020scaling,hoffmann2022training}. Yet, while model architectures and scaling laws have been extensively studied and well-documented, the methodology for transforming raw data into high-quality training corpora remains fragmented and under-theorized~\citep{penedo2024fineweb,soldaini2024dolma,wang2024mathpile,su2024nemotron,gunasekar2023textbooks}. The field lacks a systematic framework to categorize, compare, and reason about data processing operations. This absence of a unified taxonomy forces practitioners to rely on ad-hoc experimentation, hindering reproducibility and obscuring the principled relationship between specific data transformations and downstream model capabilities.

We introduce \textbf{Data Darwinism}, a conceptual framework that treats data processing as an endless evolutionary process rather than a one-time engineering task. At its core lies a ten-level hierarchy (L0--L9) that systematically organizes data operations along multiple fundamental dimensions:

\begin{itemize*}
    \item \textbf{From Selection to Generation}: Lower levels (L0--L3) focus on filtering and preserving original content, while higher levels (L7--L9) transition to synthesizing entirely new environments and worlds.
    \item \textbf{From Preservation to Transformation}: Intermediate levels (L4--L6) introduce model-driven refinement that actively rewrites and enriches content while maintaining semantic fidelity.
    \item \textbf{From Human-Centric to Machine-Driven}: As data ascends the hierarchy, processing shifts from rule-based heuristics to sophisticated generative models capable of cognitive reasoning and contextual completion.
\end{itemize*}

Central to this framework is a \textbf{co-evolutionary feedback loop}: more capable models enable more sophisticated data processing techniques (e.g., using advanced large language models (LLMs) for quality assessment, content rewriting, and reasoning augmentation), which in turn produces higher-quality training data for the next generation of models. In this view, ``data quality'' is not a static attribute but a moving target that evolves with the expanding frontier of model capabilities.

To operationalize and validate Data Darwinism, we focus on the \textbf{scientific domain}---a frontier of immense conceptual density that remains largely untapped in open-source pre-training due to systemic barriers in acquisition, parsing, and learnability~\citep{taylor2022galactica,lewkowycz2022solving,lo2020s2orc,blecher2023nougat}. We implement the first six levels of our hierarchy (L0--L5), constructing \texttt{Darwin-Science}, a rigorously processed 900B-token scientific corpus spanning academic books and research papers across natural sciences, engineering, and medicine.
Our construction pipeline reveals a critical insight: \textbf{raw scientific data suffers from a severe learnability gap}. Despite their high information density, unprocessed scientific texts---even after preliminary filtering (L0--L3)---provide negligible performance gains when used for pre-training. Diagnostic experiments show that models trained on raw scientific data perform no better than baselines on both standard benchmarks and distribution-aligned evaluations. This counter-intuitive finding identifies a fundamental challenge: the high conceptual compression, implicit reasoning chains, and expert-oriented exposition characteristic of scientific literature render raw content largely opaque to language models.

To bridge this gap, we advance our processing to the intermediate levels of the Data Darwinism hierarchy:

\begin{itemize*}
    \item \textbf{L4 (Generative Refinement)}: We deploy large language models to purify learning content by systematically removing non-educational noise (metadata, navigation elements, OCR artifacts) and repairing structural fragmentation (split equations, malformed tables). This stage isolates high-value academic content while preserving semantic integrity.
    
    \item \textbf{L5 (Cognitive Completion)}: We leverage frontier LLMs to transform expert-level writing into pedagogically enriched content. This involves (1) \textbf{reasoning reconstruction}---expanding implicit logical leaps into explicit step-by-step derivations; (2) \textbf{terminological explication}---contextualizing domain-specific jargon inline rather than assuming prerequisite knowledge; and (3) \textbf{pedagogical bridging}---grounding abstract theories in concrete analogies and established concepts. This process fundamentally lowers the cognitive barrier for models to internalize complex scientific causality.
\end{itemize*}

To rigorously validate our hierarchical approach, we establish a \textbf{controlled experimental framework} that addresses a persistent methodological gap in domain-specific pre-training research: the confounding of data quality with model configuration effects. We develop \texttt{Darwin-Science-Eval}, a challenging benchmark comprising 150K expert-level questions derived from held-out scientific literature, specifically designed to assess distribution-aligned domain comprehension beyond elementary science. More critically, we train \textit{daVinci-origin-3B} and \textit{daVinci-origin-7B}---fully transparent base models trained from scratch on a carefully curated 5.37T-token corpus that deliberately excludes all scientific content. These contamination-free checkpoints serve as clean-room baselines with robust general capabilities but zero exposure to scientific domains, enabling unambiguous attribution of performance gains to data processing strategies.

Starting from these base models, we conduct 600B tokens of continued pre-training (CPT) comparing our hierarchy-processed \texttt{Darwin-Science} against a competitive baseline mixture. The results demonstrate robust and sustained efficacy:

\begin{itemize*}
    \item \textbf{Substantial Overall Gains}: \texttt{Darwin-Science} outperforms the baseline by \textbf{+2.12 points (3B)} and \textbf{+2.95 points (7B)} averaged across 20+ diverse benchmarks, with improvements amplifying to \textbf{+5.60 and +8.40 points} on our distribution-aligned \texttt{Darwin-Science-Eval} suite.
    
    \item \textbf{Hierarchy Unlocks Value}: While L0–L3 yields negligible gain, L4 and L5 achieve cumulative improvements of +0.38 and +1.36, respectively. This confirms that systematic hierarchy is essential to unlock latent data value.
    
    \item \textbf{No Saturation Signal}: Performance gains persist and even accelerate throughout the 600B-token training window with no signs of diminishing returns, indicating that our processed corpus provides superior sustained learning value at scale.
    
    \item \textbf{Model Scale Amplifies Benefits}: Larger models derive disproportionately greater value from scientific data (7B: +2.95 vs. 3B: +2.12), suggesting that model capacity is a critical determinant of data utilization for high-complexity content.
\end{itemize*}

Beyond validation, our controlled setting enables \textbf{evidence-based guidelines} for practitioners:

\begin{itemize*}
    \item \textbf{Data Composition}: A 50\% scientific content ratio optimizes the balance between domain specialization and general capabilities; internal book-to-paper ratios show high flexibility, yet including both is recommended for their complementary value.
    
    \item \textbf{Processing Strategy}: Teacher model quality directly determines cognitive completion effectiveness, with Qwen3-235B yielding +0.52 over GPT-OSS-120B.
    
    \item \textbf{Model Properties}: Extended context (32K vs. 4K) provides +0.80 advantage after sufficient adaptation; scientific data benefits persist across training stages (early 930B vs. late 4T checkpoints), validating early-stage evaluation as a compute-efficient proxy.
    
    \item \textbf{Evaluation Alignment}: Domain-matched benchmarks reveal 3$\times$ larger gains than standard evaluations, emphasizing the necessity of distribution-aligned assessment.
\end{itemize*}

In summary, this work makes three primary contributions:

\begin{enumerate*}
    \item \textbf{Conceptual Framework}: We introduce Data Darwinism, the first systematic hierarchy for categorizing and reasoning about data processing operations, establishing shared principles for the field.
    

    \item \textbf{Practical Implementation}: We construct \texttt{Darwin-Science} by operationalizing this framework on scientific literature, creating the largest open-source, hierarchically processed 900B-token scientific corpus\footnote{To benefit the academic and communities, we have open-sourced a subset of our high-quality corpus, totaling 496B tokens. This includes 82B tokens of L4-level data and 250B tokens of L5-level data, as well as an additional 164B tokens of L5-level data processed through the GPT-OSS-120B.}---and releasing the transparent \textit{daVinci-origin} base models to the community.
    
    \item \textbf{Empirical Validation}: Through rigorous controlled experiments, we demonstrate that systematic progression through the processing hierarchy is not merely beneficial but essential for unlocking the value of conceptually dense domains, and we derive actionable guidelines for data mixture, processing depth, and evaluation strategy.
\end{enumerate*}

By bringing theoretical order to the currently fragmented landscape of data engineering and providing concrete evidence of hierarchy-driven value unlocking, Data Darwinism offers both a conceptual foundation and a practical roadmap for advancing the next generation of scientific AI systems grounded in principled, co-evolutionary data-model development.

\section{Data Processing Hierarchy in Data Darwinism}

\begin{figure}
    \centering
    \includegraphics[width=1\linewidth]{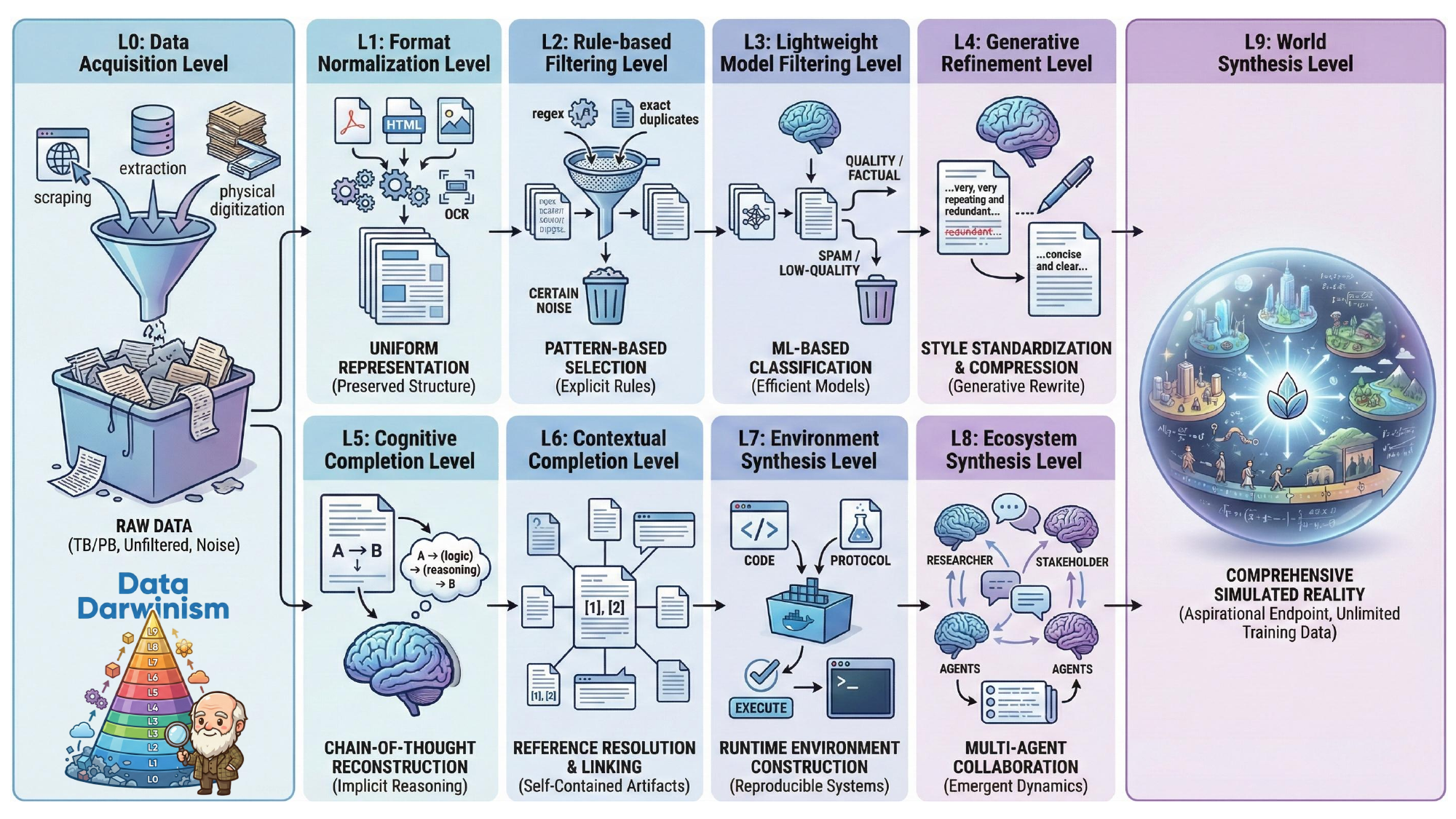}
    \caption{Overview of Data Processing Hierarchy in Data Darwinism}
    \label{fig:data_darwinism}
\end{figure}


We define a ten-level hierarchy (L0--L9) to systematically categorize data processing workflows based on their degree of transformation and value addition (Fig.~\ref{fig:data_darwinism}). This progression tracks data as it moves from initial acquisition to simulated synthesis. As data ascends through these levels, it follows a characteristic trade-off: while total volume typically decreases, the quality, information density, and structural complexity increase. This shift reflects a transition from selective filtering at lower levels to sophisticated, model-driven enrichment at higher stages, ultimately maximizing the learning value per token.

\paragraph{L0: Data Acquisition Level}
Data Acquisition (L0) represents the foundational stage where raw data is collected from diverse sources such as web scraping, database extraction, and others. This level handles the largest data volumes—typically terabytes to petabytes—in highly variable formats such as HTML, PDF, binary files, images, and video. While data quality at this stage is inherently inconsistent, containing significant noise and duplicates, L0 deliberately preserves the original information landscape to maximize downstream flexibility. The primary technical challenges center on achieving broad coverage, maintaining data provenance, and managing large-scale storage infrastructure. Computation at this level is predominantly I/O-bound, constrained by network bandwidth and storage capacity rather than GPU resources.

\paragraph{L1: Format Normalization Level}
Format Normalization (L1) transforms heterogeneous data into unified, training-ready representations, with specific operations determined by downstream task requirements. For text-based training, key operations include performing OCR on PDFs and images, parsing HTML to obtain clean content (removing markup and scripts), and transcribing audio/video. The goal is to ensure uniform processability without filtering content, while preserving structural fidelity such as document hierarchy. Computational demands vary significantly: OCR and audio transcription are GPU-intensive, while HTML parsing is relatively lightweight, though all processes are generally parallelizable. Data volume remains comparable to L0, with focus on resolving encoding issues and standardizing metadata for downstream compatibility.

\paragraph{L2: Rule-based Filtering Level}
Rule-based Filtering (L2) introduces the first stage of quality control through deterministic, pattern-based mechanisms. This level applies explicit rules to filter objectively identifiable problematic content, such as documents that are too short or excessively long, exact or near-duplicate content detected through deduplication algorithms like MinHash, garbled text and encoding errors, content in non-target languages, and text with abnormal ratios of special characters or repetitive patterns. This approach is predictable, interpretable, and highly efficient, running effectively on CPU infrastructure without requiring machine learning models. L2 achieves substantial data volume reduction while maintaining high scalability.

\paragraph{L3: Lightweight Model Filtering Level}

Lightweight Model Filtering (L3) introduces machine learning-based classification capabilities using pre-trained lightweight models such as FastText or small-scale language models to perform semantic-level judgments. Unlike L2's surface-level pattern matching, L3 understands semantic features of content, enabling tasks such as topic categorization, domain identification, quality assessment, and evaluation of educational value. L3 remains focused on filtering functionality, deciding content retention or discard without modification, while balancing model capability with computational efficiency. This layer further refines the dataset by filtering out documents that do not align with training requirements.

\paragraph{L4: Generative Refinement Level}
Generative Refinement (L4) marks a shift from selection to active, model-driven refinement using medium-to-large generative models. This level focuses on purifying content by removing extraneous noise and repairing structural or formatting defects while strictly adhering to the original content. A critical constraint is that L4 must be a faithful refiner, ensuring no external knowledge is introduced. By standardizing presentation and resolving artifacts, L4 transforms raw data into a coherent, learning-ready format without altering the underlying information.

\paragraph{L5: Cognitive Completion Level}
Cognitive Completion (L5) employs generative models to explicate the implicit reasoning and logical steps underlying existing content. Unlike L4, which transforms expression, L5 enriches data by reconstructing the "chain of thought" for mathematical derivations, scientific arguments, and instructional trajectories. Technical implementation leverages Chain-of-Thought prompting and process supervision. Quality control is complex, requiring domain-specific verification of the reasoning process rather than just factual accuracy. This enriched data is substantial for training AI systems with advanced problem-solving capabilities.

\paragraph{L6: Contextual Completion Level}

Contextual Completion (L6) expands data by integrating external references and background knowledge to resolve implicit dependencies. Recognizing that documents often cite concepts without definitions, L6 systematically retrieves and links cited sources, related work, and prerequisite definitions to create self-contained artifacts. Key operations include reference resolution and cross-referencing using semantic search and knowledge graph technologies. While this process can dramatically expand dataset size, the primary challenge lies in determining the appropriate scope to prevent information overload while ensuring comprehensive understanding.

\paragraph{L7: Environment Synthesis Level}
Environment Synthesis (L7) transcends content enrichment to construct executable, interactive environments where data objects function. Moving from static artifacts to dynamic systems, L7 synthesizes the specific runtime conditions—such as OS configurations for code (Docker/VM specifications) or experimental setups for scientific protocols—required for reproducibility. Technical implementation demands multi-modal reasoning to infer infrastructure dependencies and verify system compatibility. The goal is to generate environments that validly execute or simulate the original data, applied selectively where operational context is crucial for utilization.

\paragraph{L8: Ecosystem Synthesis Level}
Ecosystem Synthesis (L8) constructs dynamic multi-agent systems where diverse intelligent entities interact and evolve. Unlike the static environments of L7, L8 creates populated ecosystems where agents—such as simulated researchers or business stakeholders—engage in sustained collaboration, debate, and strategy. The value lies in the emergent data generated through operation: conversation logs, decision traces, and novel scenarios arising from collective activity. Implementation requires integrating language models for cognition with simulation engines, demanding high computational resources to support continuous inference across multiple adapting agents.

\paragraph{L9: World Synthesis Level}
World Synthesis (L9) represents the theoretical apex of data processing: constructing comprehensive, physically and socially coherent simulated worlds. L9 aspires to create alternative realities with internal logic—complete with physical laws, emergent civilizations, and open-ended evolution—using original data as the seed. For instance, a physics textbook might parameterize a universe's laws. While currently facing immense computational and theoretical challenges regarding scale and consistency, L9 defines the aspirational endpoint where synthetic history and intelligence emerge from deep simulation, offering essentially unlimited training data.

\section{Dataset Construction}
\label{sec:data_construction}

\begin{figure}
    \centering
    \includegraphics[width=1\linewidth]{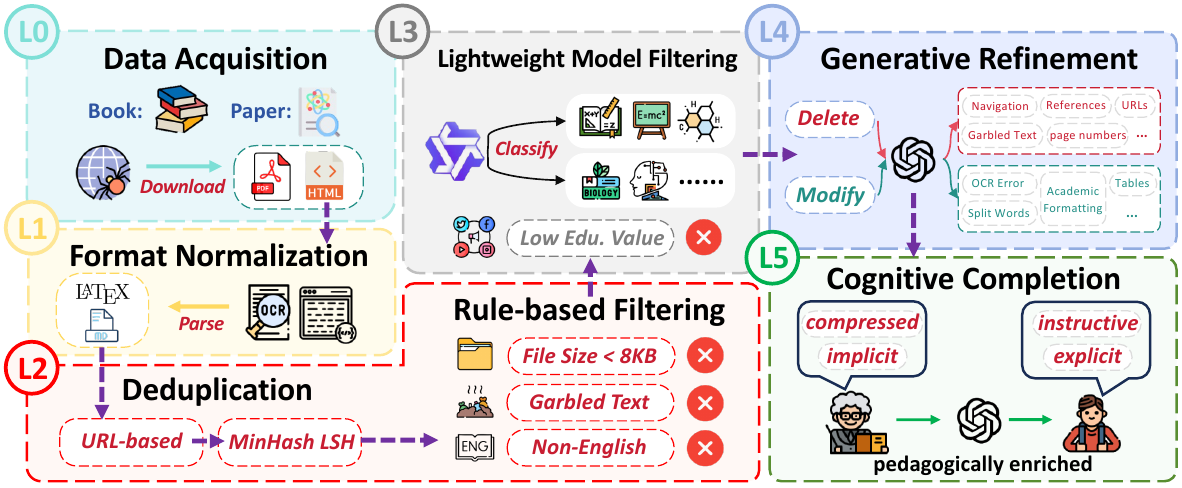}
    \caption{Overview of the dataset construction pipeline.}
    \label{fig:data_pipeline}
\end{figure}

To evaluate the Data Darwinism framework, we operationalize its hierarchy through \texttt{Darwin-Science}, a large-scale scientific corpus. This domain serves as an ideal testing ground: it remains significantly underexplored--lacking even basic large-scale open-source L0 corpora--while its high conceptual density creates steep learnability barriers that standard processing cannot breach. We implement a six-stage pipeline (L0--L5), transitioning from initial acquisition and normalization (L0--L1) to systematic filtering (L2--L3), and ultimately to generative refinement and cognitive completion (L4--L5). This structured progression demonstrates how ascending the hierarchy can systematically unlock the latent value within specialized domains, transforming fragmented raw text into a high-utility training corpus for foundation models.

\subsection{L0: Data Acquisition}
\label{Data Sources}

Our data primarily originates from publicly accessible academic resources and open-source datasets. Specifically, we collected data from the following sources:

\paragraph{Publicly accessible resources} We gathered academic books and papers from multiple publicly accessible online repositories. The book collection includes academic monographs, textbooks, and technical literature, spanning multiple disciplines. The paper collection encompasses academic papers, journal publications, and related scholarly works across various academic fields.The raw materials primarily consist of scanned PDF files.

\paragraph{Open-source dataset} From the open-source dataset TxT360  ~\cite {txt360data2024}, we selected three scholarly paper collections: (1) PubMed Central, containing biomedical and life sciences literature; (2) arXiv, comprising preprints in physics, mathematics, computer science, and related domains; and (3) S2ORC full text, encompassing diverse academic publications across multiple disciplines.


\label{Initial Processing}
\subsection{L1: Format Normalization} 
Since our work primarily focuses on training text-based models, we converted scanned PDFs into machine-readable text using \texttt{olmOCR-7B-0225-preview}, a 7B-parameter vision-language model optimized for document text extraction \citep{poznanski2025olmocr}.




\subsection{L2: Rule-based Filtering} 
\label{L2: Rule-based Filtering}


To remove redundancy, we apply MinHash~\citep{10.1007/3-540-45123-4_1} with LSH from \texttt{datatrove}~\citep{penedo2024datatrove} using parameters $(n_b, n_h) = (14, 8)$ (112 hash features per document), removing 22\% of documents. After deduplication, we apply a three-stage filtering pipeline: (1) \textbf{File Size}: we discard documents smaller than 8KB to remove spam and fragments; (2) \textbf{Garbled Text}: we filter out documents with more than 50\% garbled characters resulting from OCR errors; (3) \textbf{Language}: we retain only English documents using \texttt{fast-langdetect}.

\subsection{L3: Lightweight Model Filtering}
\label{L3: Lightweight Model Filtering}
After rule-based filtering, we annotate all documents using \texttt{EAI-Distill-0.5B}~\citep{ai2025essentialwebv1024ttokens}, a fine-tuned Qwen2.5-0.5B-Instruct model for 12-dimensional document classification covering aspects such as field of discipline classification (FDC), document type, and content quality. 

\paragraph{Non-educational Content Filtering}
Among the 12-dimensional labels generated by \texttt{EAI-Distill-0.5B}, Document Type v1 and Document Type v2 identify the content type of documents within the classification system, such as News, Personal Blog, etc. We filter out documents with no educational value based on these labels, such as Advertisement. 

\paragraph{Discipline Classification}
We further organize the retained documents into 9 major disciplines using FDC labels from \texttt{EAI-Distill-0.5B}: \textit{computer science}, \textit{medicine}, \textit{biology}, \textit{chemistry}, \textit{mathematics}, \textit{physics}, \textit{human \& social sciences}, \textit{engineering}, and \textit{other STEM fields}. The complete FDC mapping scheme is detailed in Appendix~\ref{app:fdc-mapping}.

\paragraph{Book-Paper Classification}
Finally, since books and papers exhibit different learnability characteristics that require different downstream processing, we classify all documents into \textit{book} and \textit{paper} categories. For data sources with explicit type metadata (e.g., arXiv papers, published books), we directly use the provided labels; for ambiguous cases, we employ \texttt{Qwen2.5-7B-Instruct}~\citep{qwen2.5} to determine whether each document is a book or a paper. The classification methodology and processing differences between the two categories are elaborated in Appendix~\ref{app:book-paper-split}.

\subsection{L4: Generative Refinement}
\label{sec:structural_denoising}

This stage addresses the \textit{learnability gap} identified in our diagnostic experiments. While preliminary filtering (L2--L3) ensures document-level relevance, scientific texts, particularly those derived from scanned sources, often contain persistent structural noise, such as reference lists, malformed equations, and OCR-induced errors. These elements act as distractions that disrupt the model's focus on core scientific logic. As a faithful refiner, L4 moves beyond discarding documents to actively purifying the internal learning signal.

\paragraph{Approach Design}
To systematically address these quality issues, we develop an LLM-based refinement approach guided by an empirical analysis. 
Our strategy centers on two core principles to minimize extraneous content while preserving high-value academic integrity, such as technical notation and educational materials (full prompt in Appendix~\ref{app:l3_prompt}):
\textbf{Deletion:} Removing minimal-value content such as structural elements (table of contents, references, headers/footers), non-academic artifacts (placeholders, URLs, advertisements), OCR errors (garbled text, encoding anomalies), and scanning duplications.
\textbf{Modification:} Repairing formatting defects without altering semantics, such as merging fragmented text and restoring damaged formulas or tables.

\paragraph{Implementation}
We apply this refinement pipeline to the entire OCR-processed corpus. 
Documents are segmented into 1,024 character chunks ($\approx$256 token windows) and processed independently at scale using \texttt{GPT-OSS-120B}~\citep{openai2025gptoss120bgptoss20bmodel}, selected for its optimal balance of accuracy and throughput (see Appendix~\ref{app:l3_evaluation_and_model_section}). 
This granularity balances refinement fidelity with context preservation: it is small enough to ensure strict adherence to refinement principles, yet large enough to maintain textual coherence.
The process results in a 20\% reduction in corpus volume; additional implementation details and examples are provided in Appendices~\ref{app:l3_implementation_details} and~\ref{app:l3_examples}.

\subsection{L5: Cognitive Completion}

While the generative refinement described in Sec.~\ref{sec:structural_denoising} ensures data cleanliness, research corpus is typically written in an "Expert-to-Expert" paradigm, characterized by high information compression, implicit reasoning steps, and heavy reliance on assumed background knowledge. For a pre-training model, this creates a \textit{understanding barrier}: the model encounters conclusions without witnessing the derivation process, leading to inefficient internalization of logic. 

\paragraph{Approach Design} To bridge this gap, we introduce a \textbf{Cognitive Completion} strategy. We employ a pipeline designed to make implicit reasoning explicit. Specifically, the augmentation targets three key dimensions: (1) \textbf{Reasoning Reconstruction:} Expanding logical leaps (e.g., "it follows that") into step-by-step derivations, allowing the model to trace the causality between assumptions and conclusions.
(2) \textbf{Terminological Explication:} Contextualizing domain-specific jargon and variable definitions within the narrative flow rather than assuming prior mastery.
(3) \textbf{Pedagogical Bridging:} We ground abstract concepts in established knowledge through intuitive analogies. This involves introducing contextual bridges that link complex, isolated theoretical constructs to concrete physical examples, facilitating better concept association.

\paragraph{Implementation}
Given the high conceptual density of research papers and the substantial computational cost of generative rewriting, we apply this augmentation exclusively to paper rather than books. To ensure tractable processing while maintaining narrative consistency, we segment documents into 1,024 token windows (a larger window size compared to Sec.~\ref{sec:structural_denoising}). The rewriting process is executed by \texttt{Qwen3-235B-A22B-Instruct}~\cite{yang2025qwen3}, guided by a structured prompt (see Appendix~\ref{L5 Processing Prompt}) designed to strictly enforce the dimensions described above. The rewrite model selection is based on a preliminary study using an \textit{LLM-as-a-Judge} framework (detailed in Appendix~\ref{sec:l4_eval_iter}).

\subsection{Data Portrait}

Before finalizing the dataset, we conducted a decontamination process to mitigate benchmark leakage. Specifically, we checked the overlap between our corpus and several widely used downstream benchmarks for evaluating LLM performance, including GSM8K~\cite{cobbe2021gsm8k}, MATH~\cite{hendrycksmath2021}, and MMLU~\cite{hendryckstest2021}. We concatenated problems and solutions as complete samples, performed exact 20-gram matching, and excluded any contaminated documents, which removed approximately 0.03\% of the data.


The final \texttt{Darwin-Science} comprises 50M documents totaling approximately 900B tokens, with broad coverage across natural sciences, engineering, and social sciences. Detailed statistics are in Tab.~\ref{tab:data-stats} and Appendix~\ref{app:scipedia-distribution}. We also construct Darwin-Science-Raw, containing 601B tokens (321B from books, 280B from papers) of original OCR-extracted text. We believe this high-quality dataset will provide a valuable resource for the community.

\begin{table}[h]
\centering
\caption{Dataset statistics across categories}
\small 
\begin{tabular}{lccc}
\toprule
\textbf{Category} & \textbf{Samples} & \textbf{Tokens} & \textbf{Avg.} \\
 & \textbf{(M)} & \textbf{(B)} & \textbf{Toks/Sample} \\
\midrule
\rowcolor{green!10}
\textbf{Book} & 2.98 & 251.5 & 84396 \\
\hspace{1em}\textit{L4} & 2.98 & 251.5 & 84396 \\
\rowcolor{green!10}
\textbf{Paper} & 47.81 & 655 & 13700 \\
\hspace{1em}\textit{L4} & 26.31 & 215 & 8172 \\
\hspace{1em}\textit{L5} & 21.50 & 440 & 20465 \\
\midrule
\rowcolor{yellow!30}
\textbf{Total} & 50.79 & 906.5 & 17848 \\
\bottomrule
\end{tabular}
\label{tab:data-stats}
\end{table}

\section{Evaluation of Darwin-Science}
\label{sec:scipedia_eval}


Existing benchmarks target elementary science and lack the depth to capture the specialized knowledge enhanced by \texttt{Darwin-Science}. To address this, we introduce \texttt{Darwin-Science-Eval}, an academic benchmark designed to evaluate this specialized nature. We generate seven-option multiple-choice questions via a three-stage pipeline.

\paragraph{Q\&A Generation} First, we intelligently segment original documents into chunks of 4096 tokens, ensuring relative semantic completeness of each segment. Subsequently, we employ carefully designed prompts to drive the Qwen3-32B model's thinking reasoning mode, enabling the model to deeply analyze each text segment and determine whether it contains knowledge points suitable for generating evaluation questions. For segments suitable for question generation, the model further identifies the most valuable and representative knowledge points and generates high-quality multiple-choice questions accordingly .

To fundamentally ensure the correctness of questions and answers, we adopt a key constraint strategy: requiring that both the knowledge points examined in the questions and the correct answers must be directly grounded in the original text. The model only performs text refinement and reorganization, rather than relying on its own knowledge base for independent design.

\paragraph{Completeness Filter} The first-stage filtering focuses on examining question independence and self-containment. We require that each question must be independently assessable without relying on any external information beyond the question text, nor should it contain referential expressions pointing to external content. We employ the Qwen3-32B model, inputting only the question itself for independence assessment, ensuring that each question functions as an independent evaluation unit.

\paragraph{Correctness Filter} Building upon completeness validation, we further implement second-stage correctness verification. In this stage, we input the original text, question, and answer together into the Qwen3-32B , requiring it to determine whether the labeled correct answer can be sufficiently supported by the original text. Only questions whose answers can be clearly verified against the original text are retained. Through this dual filtering mechanism of independence and correctness, we significantly enhance the quality and reliability of the final benchmark.

Following this pipeline, we construct \texttt{Darwin-Science-Eval}, comprising 140K questions from books and 10K questions from papers, all sourced from documents held out from the training data. To enable efficient evaluation during pretraining, we sample 1,500 questions from both book and paper to form two test sets: \texttt{Darwin-Science-Eval-Book} and \texttt{Darwin-Science-Eval-Paper}.

\begin{figure}
\centering
\includegraphics[width=0.9\linewidth]{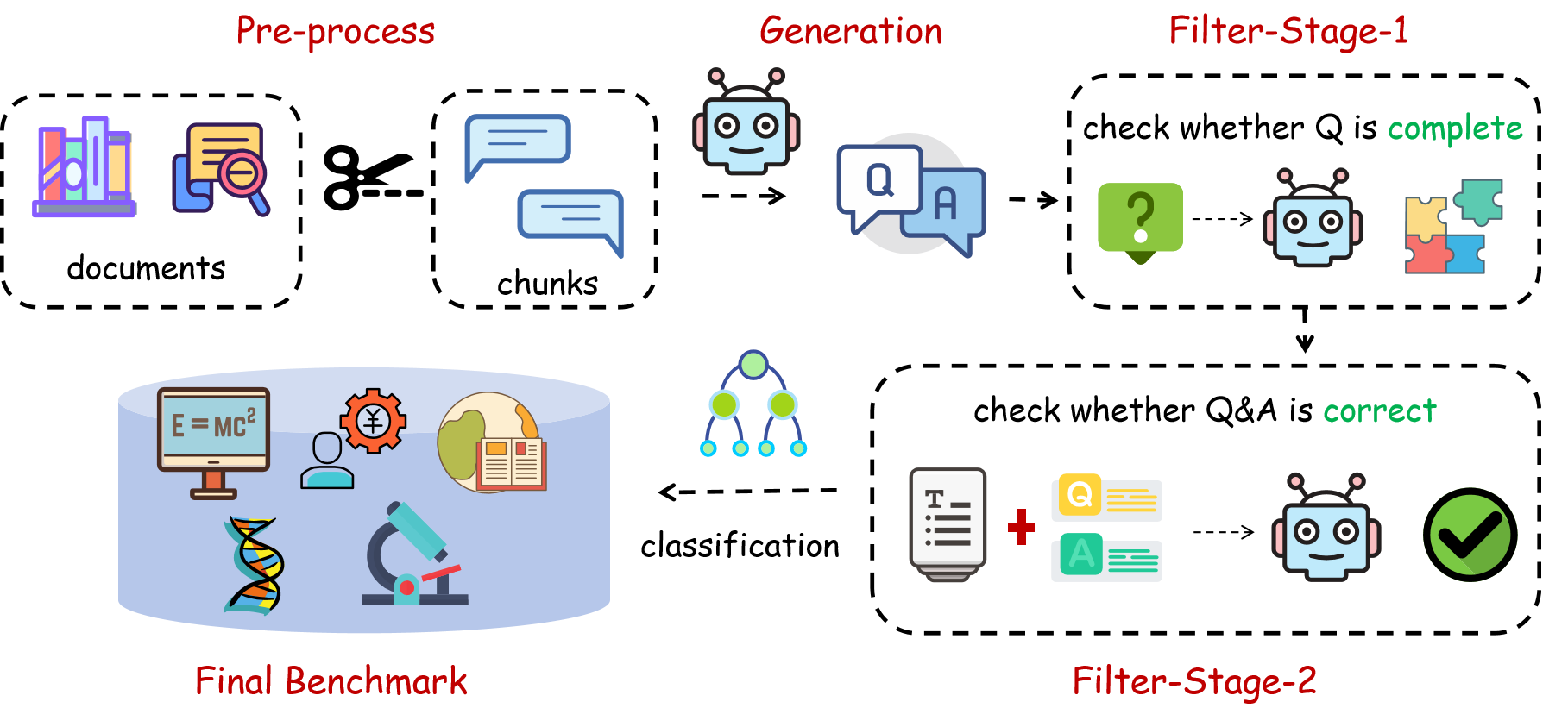}
\caption{\label{fig:bench-generate_pipeline} Construction pipeline of our benchmark}
\end{figure}

\section{Foundation Model Training}
We train \textit{daVinci-origin-3B} and \textit{daVinci-origin-7B} from scratch to provide fully transparent foundation models, avoiding the ``black-box" of off-the-shelf alternatives. By strictly excluding scientific content during pre-training, we establish contamination-free base models with robust foundational capabilities while remaining unexposed to the scientific domain. These models serve as a controllable research foundation with a fully disclosed data recipe, offering the research community a clean-room environment for studying domain-specific knowledge acquisition and data influence.

\label{sec:pretraining}
\subsection{Pretraining Dataset}
\label{sec:pretraining_dataset}

Our foundation model training dataset  consists of three parts: CC, Math, and Code, totaling 5.37T. The specific composition can be seen in Table~\ref{tab:dataset}



\paragraph{CC.}Massive web data accounts for a significant portion of pretraining. We selected the non-synthetic subset of the Nemotron-CC ~\cite{su2024nemotron}. To avoid introducing additional confounding factors, we only used the real data portion of Nemotron-CC, i.e., 4.4T tokens. Then we use same discipline classification in ~\ref{L3: Lightweight Model Filtering}. After accounting for losses during processing, our final CC dataset contains approximately 4.28T tokens.

\paragraph{Math.}To enhance the model's scientific reasoning capabilities, we specifically collected two high-quality mathematical pretraining datasets: \textit{MegaMath} ~\cite{zhou2025megamath} and \textit{Nemotron-CC-Math-v1} \cite{karimi2025nemotroncc}. \textit{MegaMath} is currently the largest open-source English mathematics corpus. We selected three subsets: MegaMath-Web(264B tokens), MegaMath-Web-Pro(15B tokens), MegaMath-Synth-Code(7B tokens). Our other mathematical data source is \textit{Nemotron-CC-Math-v1}, a high-quality mathematical pretraining dataset extracted from Common Crawl. we utilized three datasets in total: Nemotron-CC-Math-v1-3 (81B tokens), Nemotron-CC-Math-v1-4+ (52B tokens), and Nemotron-CC-Math-v1-4+-MIND (74B tokens).

\paragraph{Code.} Our code dataset is derived from three sources: self-crawled GitHub repositories, Nemotron-Pretraining-Code-v1 ~\cite{nvidia2025nvidianemotronnano2}, and txt360-stack-exchange. For the self-crawled GitHub data, we filtered out repositories with fewer than 10 stars to ensure basic quality and maintenance activity. Next, we organized all the source files and applied the OpenCoder filtering method to remove low-quality or non-informative code files. Through this process, we obtained approximately 187B tokens of high-quality code data. In addition to our self-crawled GitHub data, we incorporated Nemotron-Pretraining-Code-v1 as a supplement. We crawled additional original code based on the provided metadata, and then deduplicated it against our own crawled data, and ultimately obtained 220B tokens. Furthermore, this dataset also includes large-scale natural language-code paired data constructed via LLM across 11 programming languages, namely the Synthetic-Code subset. We also utilized all synthetic data from this dataset (171B tokens). Additionally, to enrich code-related question-answer data, we incorporated the txt360-stack-exchange subset, which aggregates question-answer data from the Stack Exchange platform, totaling approximately 20B tokens.

\begin{table}[htbp]
\centering
\caption{Foundation Model Pre-training Dataset Composition}
\begin{tabular}{lr}
\toprule
\textbf{Dataset} & \textbf{Tokens} \\
\midrule
\rowcolor{blue!10}
\textbf{Common Crawl} & \textbf{4.28T} \\
\hspace{1em}Nemotron-CC (actual) & 4.28T \\
\midrule
\rowcolor{blue!10}
\textbf{Math} & \textbf{493B} \\
\hspace{1em}MegaMath & 286B \\
\hspace{2em}\textit{MegaMath-Web} & 264B \\
\hspace{2em}\textit{MegaMath-Web-Pro} & 15B \\
\hspace{2em}\textit{MegaMath-Synth-Code} & 7B \\
\hspace{1em} Nemotron-CC-Math-v1 & 207B \\
\hspace{2em}\textit{Nemotron-CC-Math-v1-3} & 81B \\
\hspace{2em}\textit{Nemotron-CC-Math-v1-4+} & 52B \\
\hspace{2em}\textit{Nemotron-CC-Math-v1-4+-MIND} & 74B \\
\midrule
\rowcolor{blue!10}
\textbf{Code} & \textbf{598B} \\
\hspace{1em}Self-crawled GitHub (star$>$5) & 187B \\
\hspace{1em}Nemotron-Pretraining-Code-v1 & 391B \\
\hspace{2em}\textit{Original} & 220B \\
\hspace{2em}\textit{Synthetic-Code} & 171B \\
\hspace{1em}txt360-stack-exchange & 20B \\
\midrule
\rowcolor{yellow!30}
\textbf{Total} & \textbf{5.37T} \\
\bottomrule
\end{tabular}
\label{tab:dataset}
\end{table}

\subsection{Pretraining Configuration}
\label{sec:pretraining_configuration}
\paragraph{Model Architecture.}
Our main experiments utilize a 3B parameter base model following the Qwen2.5 architecture~\citep{qwen2.5}. The model employs the Qwen’s tokenizer~\citep{bai2023qwen} with a vocabulary size of 151,643 tokens, a context length of 4,096 tokens, and Rotary Position Embeddings (RoPE,~\citealt{su2021roformer}) with a base frequency of 10,000.

\paragraph{Optimization Setup.}
We train all models using the AdamW optimizer~\citep{loshchilov2019decoupled} with $\beta_1=0.9$, $\beta_2=0.95$, and $\epsilon=1\mathrm{e}{-8}$. The learning rate schedule incorporates a 2,000-step linear warmup phase followed by a constant peak learning rate of $3\mathrm{e}{-4}$ throughout the remaining pretraining phase. All models use a micro-batch size of 4.

\paragraph{Training Schedule.}
We employ a progressive global batch size (GBS) scaling strategy:
\begin{itemize}
    \item \textbf{Stage 1}: GBS=1,024 for 70,000 steps ($\sim$293.6B tokens)
    \item \textbf{Stage 2}: GBS=2,048 for 40,000 steps ($\sim$335.5B tokens)
    \item \textbf{Stage 3}: GBS=4,096 for the remaining steps
\end{itemize}

The final stage varies by model configuration to achieve target token counts. This results in two 3B model variants: \textit{daVinci-origin-3B} (18,000 steps in Stage 3, $\sim$302B tokens, 930B total), and \textit{daVinci-origin-3B}$_{4\text{T}}$ (200,000 steps, $\sim$3.36T tokens, 4T total). The 7B model follows identical training recipes at the 930B token scale, denoted as \textit{daVinci-origin-7B}. All experiments are conducted using the NVIDIA NeMo framework~\citep{nemo2024}.

\paragraph{Data Mixture.} Following the data composition strategy of~\citet{allal2025smollm2,olmo20242}, where Common Crawl dominates the pretraining mixture, we adopt a sampling ratio of 80.2\% CC, 11.2\% Code, and 8.5\% Math.

\subsection{Evaluation}
\label{sec:pretraining_evaluation}

To robustly assess the effectiveness of our curated data, we conduct extensive evaluations across a wide range of mainstream benchmarks designed for large language models. Our evaluation emphasizes both general-purpose reasoning and science-oriented problem-solving abilities, complemented by two newly constructed benchmarks that specifically target complex, research-level comprehension tasks.

\paragraph{General Capabilities} To evaluate general reasoning and knowledge recall, we employ BBH (3-shot)~\citep{suzgun2022challenging}, ARC-Easy and ARC-Challenge (0-shot)~\citep{clark2018think}, MMLU (5-shot)~\citep{hendrycks2020measuring}, MMLU-Pro (5-shot)~\citep{wang2024mmlu}, DROP (5-shot)~\citep{dua2019drop}, OpenBookQA (5-shot)~\citep{OpenBookQA2018}, and PIQA (0-shot)~\citep{bisk2020piqa}.

\paragraph{Scientific Capabilities} We examine scientific domain performance using GSM-8K (8-shot)~\citep{cobbe2021training}, MATH (4-shot)~\citep{hendrycks2021measuring}, GPQA-Main (5-shot)~\citep{rein2024gpqa}, SuperGPQA (5-shot)~\citep{du2025supergpqa}, MMLU-STEM (5-shot)~\citep{hendrycks2020measuring}, MMLU-Pro-STEM (5-shot)~\citep{wang2024mmlu}, SciBench (4-shot)~\citep{wang2023scibench}, OlympicArena-MC (4-shot)~\citep{huang2024olympicarena}, MedQA (0-shot)~\citep{jin2021disease}, MedMCQA (0-shot)~\citep{pal2022medmcqa}, and PubMedQA (0-shot)~\citep{jin2019pubmedqa}.

\paragraph{Our Curated Benchmarks.}
As mentioned in Section~\ref{sec:scipedia_eval}, to address the gap in evaluating comprehension over advanced, research-level scientific materials, we further curated two multiple-choice benchmarks, \textbf{BookQA} and \textbf{PaperQA}. These datasets are designed to test deep scientific reasoning and conceptual integration derived from academic books and peer-reviewed literature.

Since the evaluated models are \textit{base checkpoints}—i.e., models not aligned or fine-tuned through post-training—we adopted both \textit{few-shot prompting} and \textit{perplexity-based} evaluation strategies to better reflect intrinsic model capability. Concretely, we used perplexity-based evaluation for \textbf{ARC-Easy}, \textbf{ARC-Challenge}, \textbf{MMLU}, \textbf{OpenBookQA}, \textbf{PIQA}, \textbf{GPQA-Main}, and \textbf{MMLU-STEM}, while generative evaluation was applied to the remaining benchmarks, particularly those requiring complex reasoning chains or CoT (Chain-of-Thought) generation. All evaluations were implemented using a slightly modified version of the \texttt{lm-evaluation-harness}~\citep{eval-harness} framework, with inference conducted under \textit{greedy decoding} settings for consistency across experiments.

\paragraph{Results}
Table~\ref{tab:pretraining_results} presents the evaluation results of our pretrained models at different training stages. We report performance for \textit{daVinci-origin-3B} at 930B tokens, \textit{daVinci-origin-3B}$_{4\text{T}}$ at 4T tokens, and \textit{daVinci-origin-7B} at 930B tokens across all benchmark categories. These pretrained models serve as capable starting points for our subsequent experiments, providing basic checkpoints with established general reasoning and scientific capabilities for investigating the impact of scientific data integration.

\begin{table}[!htp]\centering
\caption{Evaluation results of pretrained models at different training stages and scales. Abb.: G-8K (GSM-8K), SupG (SuperGPQA), M-S (MMLU-STEM), MP-S (MMLU-Pro-STEM), SciB (SciBench), Oly-MC (OlympicArena-MC), MQA (MedQA), MMCQA (MedMCQA), PMQA (PubMedQA), SiBo (Darwin-Science-Eval-book), SiPa (Darwin-Science-Eval-Paper), ARCE (ARC-Easy), ARCC (ARC-Challenge), MP (MMLU-Pro), OBQA (OpenBookQA), AVG-G (Average General), AVG-S (Average Science), Avg-D (Average In-Domain), Avg-A (Average All).}\label{tab:pretraining_results}
\resizebox{\textwidth}{!}{ 
\begin{tabular}{lrrrrrrrrrrrrrr}\toprule
&\multicolumn{11}{c}{Scientific Tasks} &\multicolumn{2}{c}{In-Domain Tasks} \\\cmidrule(r{1em}){2-12}\cmidrule(l{0.5em}){13-14}
&G-8K &MATH &GPQA &SupG &M-S &MP-S &SciB &Oly-MC &MQA &MMCQA &PMQA &SiBo &SiPa \\\midrule
\textit{daVinci-origin-3B} &20.02 &11.00 &23.88 &8.44 &34.92 &10.09 &3.34 &19.18 &30.95 &32.92 &66.00 &23.27 &19.00 \\
\textit{daVinci-origin-3B}$_{4\text{T}}$ &27.29 &12.60 &27.68 &11.31 &39.23 &12.96 &3.34 &27.07 &30.16 &34.26 &72.20 &32.60 &26.33 \\
\textit{daVinci-origin-7B} &35.41 &17.20 &24.33 &10.81 &41.33 &15.62 &3.92 &26.27 &34.64 &33.16 &73.20 &37.33 &32.93 \\\toprule
&\multicolumn{8}{c}{General Tasks} &\multicolumn{4}{c}{Average} & \\\cmidrule(r{1em}){2-9}\cmidrule(l{0.5em}){10-13}
&BBH &ARCE &ARCC &MMLU &MP &DROP &OBQA &PIQA &AVG-G &AVG-S &Avg-D &Avg-A & \\\midrule
\textit{daVinci-origin-3B} &32.31 &65.49 &36.52 &40.48 &11.20 &27.04 &38.80 &78.45 &41.29 &23.70 &21.13 &30.16 & \\
\textit{daVinci-origin-3B}$_{4\text{T}}$ &33.47 &68.64 &41.38 &45.89 &13.70 &29.64 &40.80 &77.86 &43.92 &21.85 &29.46 &33.73 & \\
\textit{daVinci-origin-7B} &36.78 &70.88 &42.49 &48.90 &18.10 &30.50 &43.80 &78.29 &46.22 &28.72 &35.13 &35.99 & \\
\bottomrule
\end{tabular}
}
\end{table}

\section{Experiments}
Leveraging the aforementioned testing ground and base models, we implement a controlled setting to evaluate our scientific corpus, specifically focusing on the impact of its hierarchical refinement. Through comparative CPT, we quantify performance gains across model scales, demonstrating how ascending the Data Darwinism hierarchy--from preliminary processing to model-driven enrichment--is essential to unlock the latent value of scientific data.

\subsection{Experimental Setup}
\label{sec:exp_setup}


\paragraph{Training Configurations}
To isolate the effect of scientific content, we compare two training configurations. Both involve 600B tokens of CPT starting from our in-house \textit{daVinci-origin-3B/7B} model, trained from scratch on a 5.37T science-free corpus (930B token checkpoint) to ensure no prior exposure to scientific domains.

\begin{itemize}
    \item \textbf{Baseline} utilizes the original pretraining mixture (80.2\% CommonCrawl, 11.2\% Code, 8.5\% Math). 
    \item \textbf{Sci-Mix} mixes 50\% of our hierarchy-processed scientific corpus (books:papers = 1:2) with 50\% baseline mixture.\footnote{The 50\% scientific ratio and 1:2 book-paper ratio are validated in Sec.~\ref{sec:analysis}.}
\end{itemize}

\paragraph{Training Details}

All experiments utilize NVIDIA NeMo framework~\citep{nemo2024} for 600B tokens of CPT with cosine decay ($3\times10^{-4} \to 3\times10^{-5}$), sequence length 4,096, and global batch size 4,096. To ensure robustness, we report the average of the final 5 checkpoints (520B--600B, saved every 1,200 steps) and smooth learning curves using a 5-point moving average.

\subsection{Main Results}
\label{sec:main_results}

\begin{figure}[t]
    \centering
    \begin{minipage}{0.48\columnwidth}
        \centering
        \includegraphics[width=\linewidth]{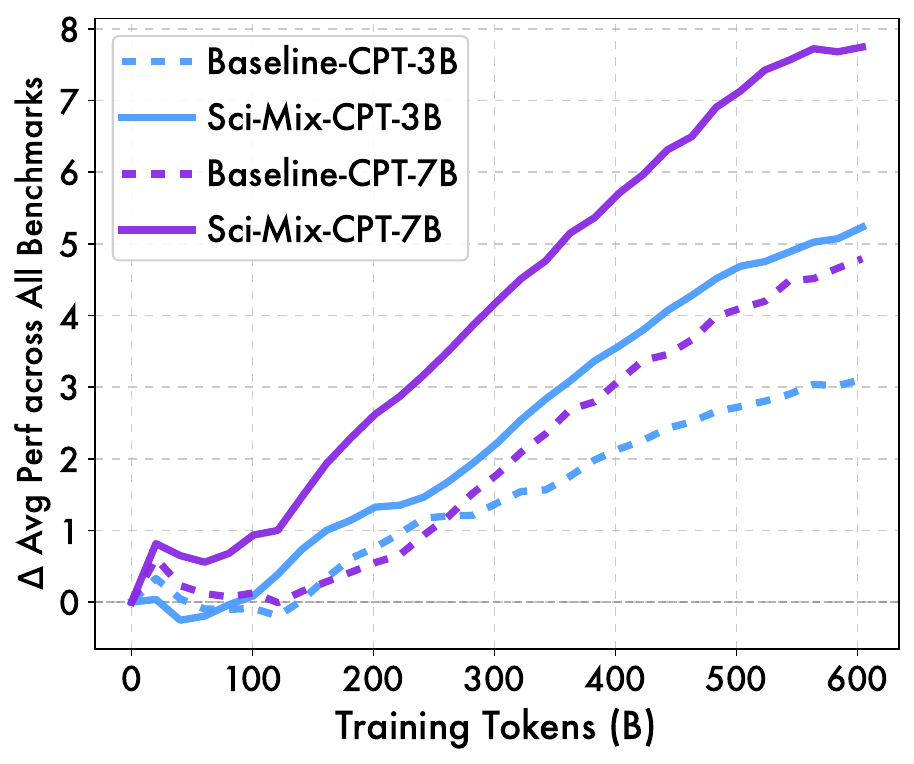}
    \end{minipage}
    \hfill 
    \begin{minipage}{0.49\columnwidth}
        \centering
        \includegraphics[width=\linewidth]{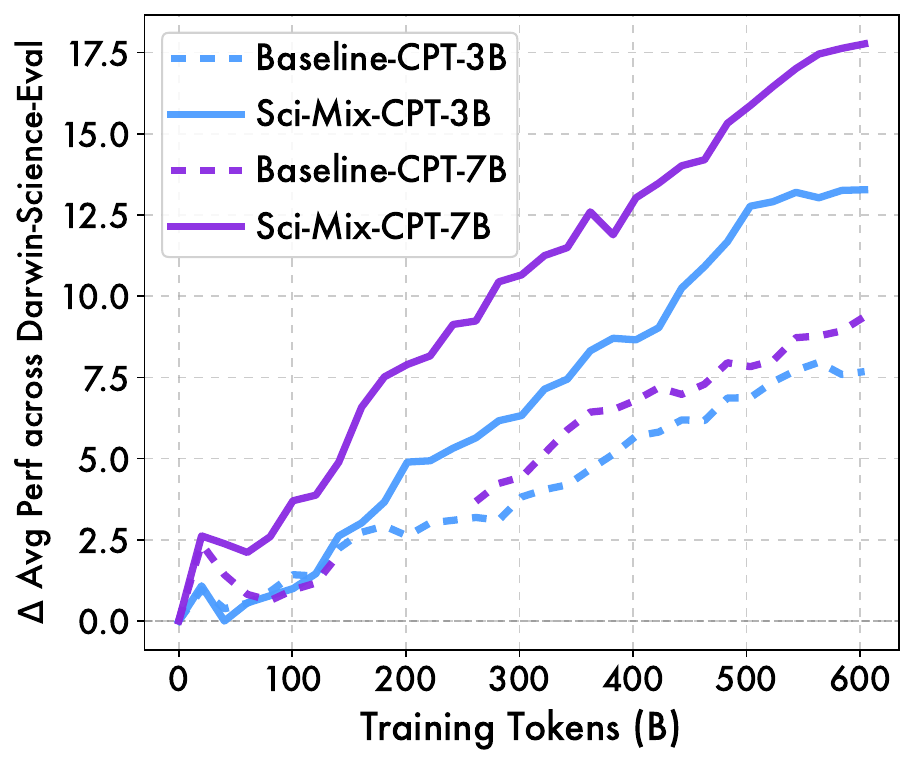}
    \end{minipage}
    
    \caption{Performance gains of \textit{daVinci-origin-3B} and \textit{daVinci-origin-7B} models. In both plots, the y-axis denotes the relative improvement over the corresponding base models.}
    \label{fig:main_results}
\end{figure}

\begin{table}[t]\centering
\caption{Performance comparison between the Baseline and Sci-Mix configurations. The Delta column denotes the improvement achieved by Sci-Mix over the Baseline.}\label{tab:main_results}

\begin{tabular}{lrrrrrrr}\toprule
&\multicolumn{3}{c}{daVinci-origin-3B} &\multicolumn{3}{c}{daVinci-origin-7B} \\\cmidrule(r{0.75em}){2-4}\cmidrule(l{0.75em}){5-7}
\textbf{} &Baseline &Sci-Mix &Delta &Baseline &Sci-Mix &Delta \\\midrule
\multicolumn{7}{c}{General Tasks} \\\midrule
BBH &33.08 &37.81 &4.73 &43.17 &49.25 &6.08 \\
ARC-Easy &66.97 &69.26 &2.29 &74.13 &74.87 &0.75 \\
ARC-Challenge &39.27 &42.05 &2.78 &49.08 &48.77 &-0.31 \\
MMLU &45.29 &48.62 &3.33 &53.19 &57.60 &4.41 \\
MMLU-Pro &13.42 &16.94 &3.52 &22.66 &27.36 &4.70 \\
DROP &29.61 &31.44 &1.82 &35.70 &37.57 &1.87 \\
OpenBookQA &42.12 &41.28 &-0.84 &45.00 &46.28 &1.28 \\
PIQA &77.45 &77.80 &0.35 &79.79 &79.72 &-0.08 \\\midrule
\multicolumn{7}{c}{Scientific Tasks} \\\midrule
GSM-8K &27.90 &29.42 &1.52 &45.97 &48.37 &2.40 \\
MATH &12.60 &12.40 &-0.20 &20.68 &20.44 &-0.24 \\
GPQA &25.80 &26.07 &0.27 &28.66 &27.28 &-1.38 \\
SupGPQA &12.11 &13.90 &1.79 &15.09 &17.34 &2.25 \\
MMLU-STEM &40.22 &39.89 &-0.33 &46.30 &50.16 &3.85 \\
MMLU-Pro-STEM &13.41 &15.67 &2.26 &20.72 &25.12 &4.40 \\
SciBench &3.84 &3.72 &-0.12 &6.51 &7.35 &0.84 \\
OlympicArena-MC &24.05 &25.72 &1.67 &30.04 &32.13 &2.09 \\
MedQA &31.11 &33.61 &2.50 &38.76 &45.78 &7.03 \\
MedMCQA &33.42 &34.53 &1.11 &37.20 &41.42 &4.22 \\
PubMedQA &69.28 &74.24 &4.96 &74.88 &75.92 &1.04 \\\midrule
\multicolumn{7}{c}{In-Domain Tasks} \\\midrule
ScienPedia-Eval-Book &30.77 &36.31 &5.53 &47.44 &53.60 &6.16 \\
ScienPedia-Eval-Paper &26.85 &32.52 &5.66 &41.56 &52.20 &10.64 \\\midrule
\multicolumn{7}{c}{Average} \\\midrule
Avg-General &45.05 &46.23 &1.18 &51.29 &52.64 &1.35 \\
Avg-Science &26.70 &28.11 &1.40 &33.17 &37.53 &4.36 \\
Avg-In-Domain &28.81 &34.41 &5.60 &44.50 &52.90 &8.40 \\
Avg-All &33.27 &35.39 &2.12 &40.79 &43.74 &2.95 \\
\bottomrule
\end{tabular}
\end{table}

\begin{figure}[t]
    \centering
    \includegraphics[width=0.49\columnwidth]{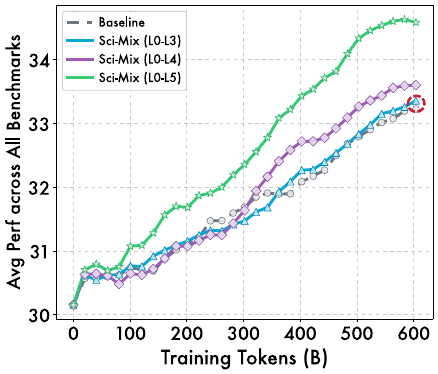}
    \hfill
    \includegraphics[width=0.49\columnwidth]{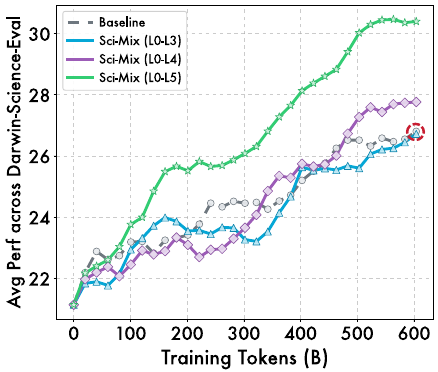}
    
    \caption{Comparison of training effectiveness across different data processing strategies.}
    \label{fig:data-processing-comparison}
\end{figure}

The quantitative results for \textit{daVinci-origin-3B} and \textit{daVinci-origin-7B} are summarized in Tab.~\ref{tab:main_results} and Fig.~\ref{fig:main_results}. 
Overall, scientific data refined through our systematic hierarchy (L0-L5)---spanning preliminary processing to model-driven enrichment---yields consistent and substantial performance improvements. We highlight four core findings:

\paragraph{Finding 1: Low-Level Processing (L0-L3) Fails to Bridge the Learnability Gap.} A critical observation in our experiments is that simply increasing the volume of scientific data does not guarantee intelligence gains. As shown in Fig.~\ref{fig:data-processing-comparison}, training on Raw Scientific Data (L0-L3), consisting of OCR-extracted text with rule-based and lightweight model-based filtering, yields negligible improvements over the Baseline, even on distribution-aligned benchmarks like \texttt{Darwin-Science-Eval}. This identifies a \textit{Learnability Gap}: despite its high conceptual density, raw scientific text remains opaque to the model, necessitating the higher-level processing defined in our hierarchy to unlock the latent value of scientific data.

\paragraph{Finding 2: Processing Hierarchy Unlocks Sustained Learning Value.} 
To realize the full potential of scientific data, systematic movement up the hierarchy is essential. By comparing processing levels in Fig.~\ref{fig:data-processing-comparison}: 
\begin{itemize}
    \item \textbf{Generative Refinement (L0--L4)}: While low-level processing (L0--L3) shows negligible results, advancing to L4 provides the first clear improvement with a cumulative gain of +0.38 points. This confirms that purifying content and repairing format defects are necessary to begin unlocking data value.
    \item \textbf{Cognitive Completion (L0--L5)}: The most significant leap occurs at the highest depth, where total gains reach \textbf{+1.36 points}. This stage drives performance by making explicit the implicit reasoning paths and intellectual scaffolding that experts often leave unstated. 
\end{itemize} 
Overall, incorporating this hierarchical pipeline yields robust gains, with \textit{daVinci-origin-3B} and \textit{daVinci-origin-7B} improving by \textbf{+2.12} and \textbf{+2.95} points on average. Critically, \textit{the advantage over the baseline grows throughout the 600B token window with no sign of saturation} (Fig.~\ref{fig:main_results}), indicating that our L0--L5 processing produces high-quality content that provides superior sustained learning value even at extended scales.

\paragraph{Finding 3: Model Capacity Amplifies Data Value.} A clear scaling pattern emerges: \textit{larger models derive greater benefits from scientific data}. \textit{daVinci-origin-7B} gains +2.95 points from scientific data compared to +2.12 for \textit{daVinci-origin-3B} (Tab.~\ref{tab:main_results}), reflecting that larger models are better equipped to capture the complex reasoning and dense domain knowledge embedded in scientific texts. While smaller models do benefit, their capacity constraints limit the extent of their learning. This suggests that for high-complexity content, model scale becomes a critical determinant of data utilization, making capacity a key consideration for effective knowledge acquisition.

\paragraph{Finding 4: Aligned Eval Reveals Hidden Gains.} \textit{Scientific data effectiveness depends heavily on the evaluation metric}. While standard benchmarks show gains of 1.76--2.38 points, aligned \texttt{Darwin-Science-Eval} yields 5.60--8.40 points---a more than threefold increase (Tab.~\ref{tab:main_results}). This stems from a distribution mismatch: standard benchmarks focus on standardized tests, whereas our training data comprises research-level content. Aligned benchmarks capture domain-specific gains that standard evaluations miss. Thus, relying solely on standard benchmarks can undervalue data sources, obscuring true gains without domain-matched evaluation.

\section{Analysis}
\label{sec:analysis}

\begin{figure*}[t] 
    \centering
    \begin{subfigure}{0.32\textwidth}
        \centering
        \includegraphics[width=\linewidth]{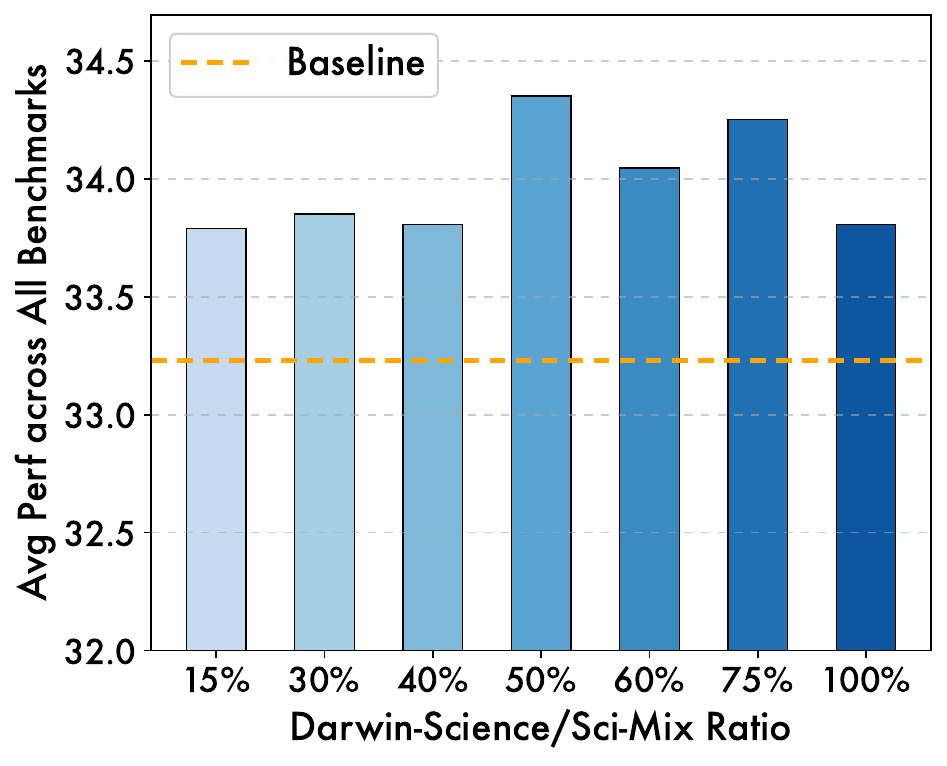}
        \caption{Effect of different overall scientific-content ratios on all benchmarks.}
        \label{fig:ratio_all_bench}
    \end{subfigure}
    \hfill
    \begin{subfigure}{0.32\textwidth}
        \centering
        \includegraphics[width=\linewidth]{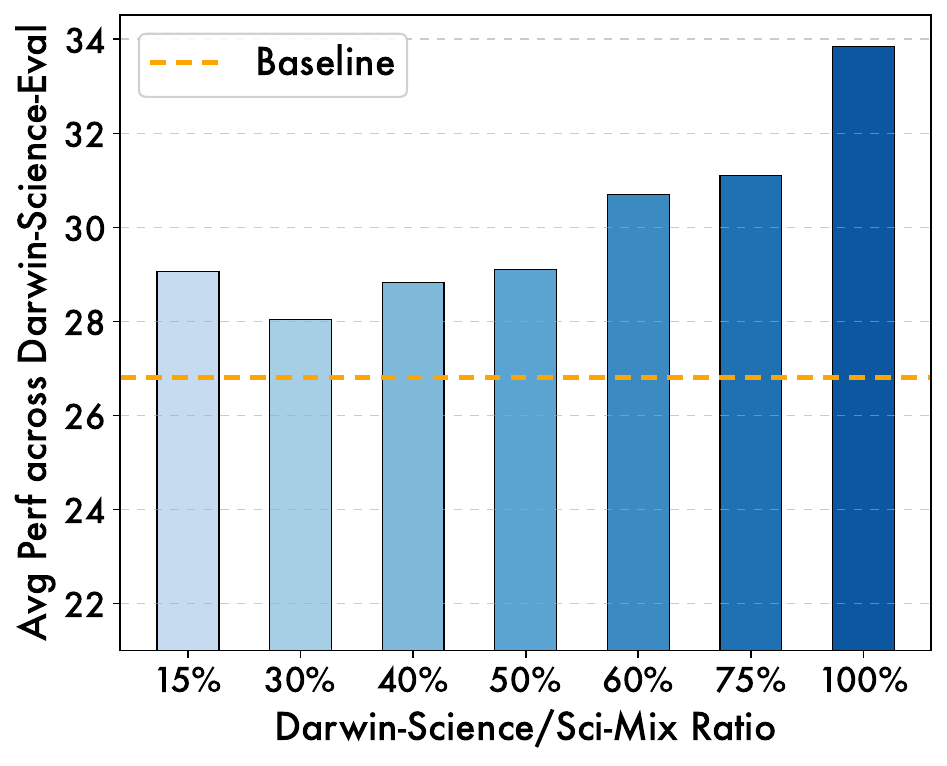}
        \caption{Effect of different overall scientific-content ratios on Darwin-Science-Eval.}
        \label{fig:ratio_scipedia_eval}
    \end{subfigure}
    \hfill
    \begin{subfigure}{0.34\textwidth}
        \centering
        \includegraphics[width=\linewidth]{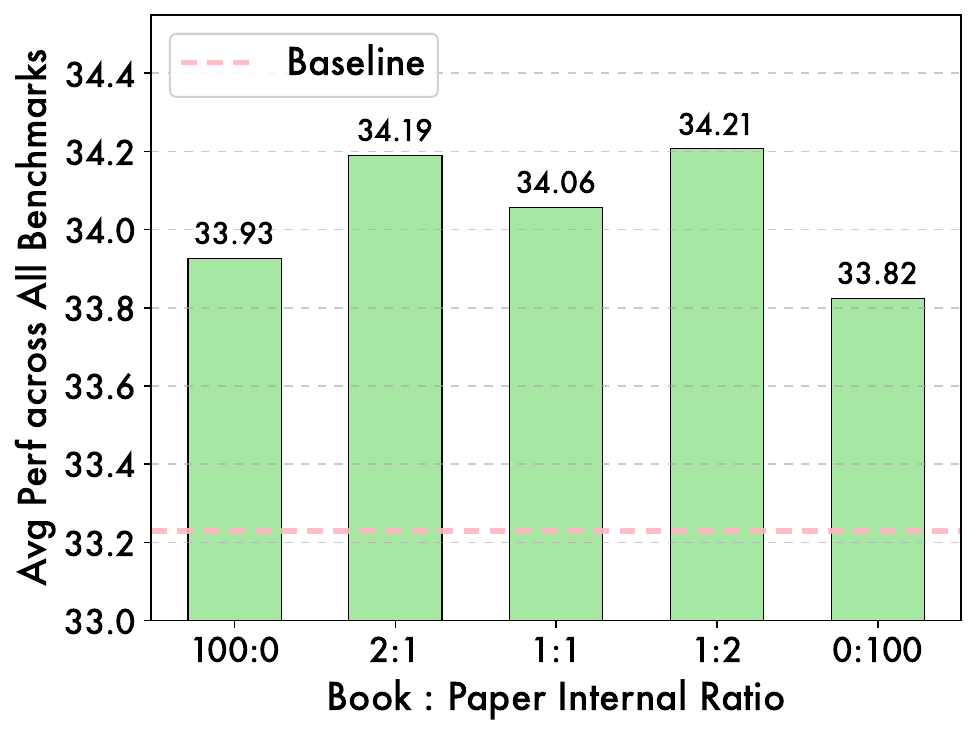}
        \caption{Effect of book-to-paper ratios within the scientific content on all benchmarks.}
        \label{fig:internal_ratio}
    \end{subfigure}

    \caption{Data-centric analysis of data mixture ratios.}
    \label{fig:scientific_content_ratio_analysis}
\end{figure*}

\begin{figure*}[t]
    \centering
    \begin{subfigure}{0.30\textwidth}
        \centering
        \includegraphics[width=\linewidth]{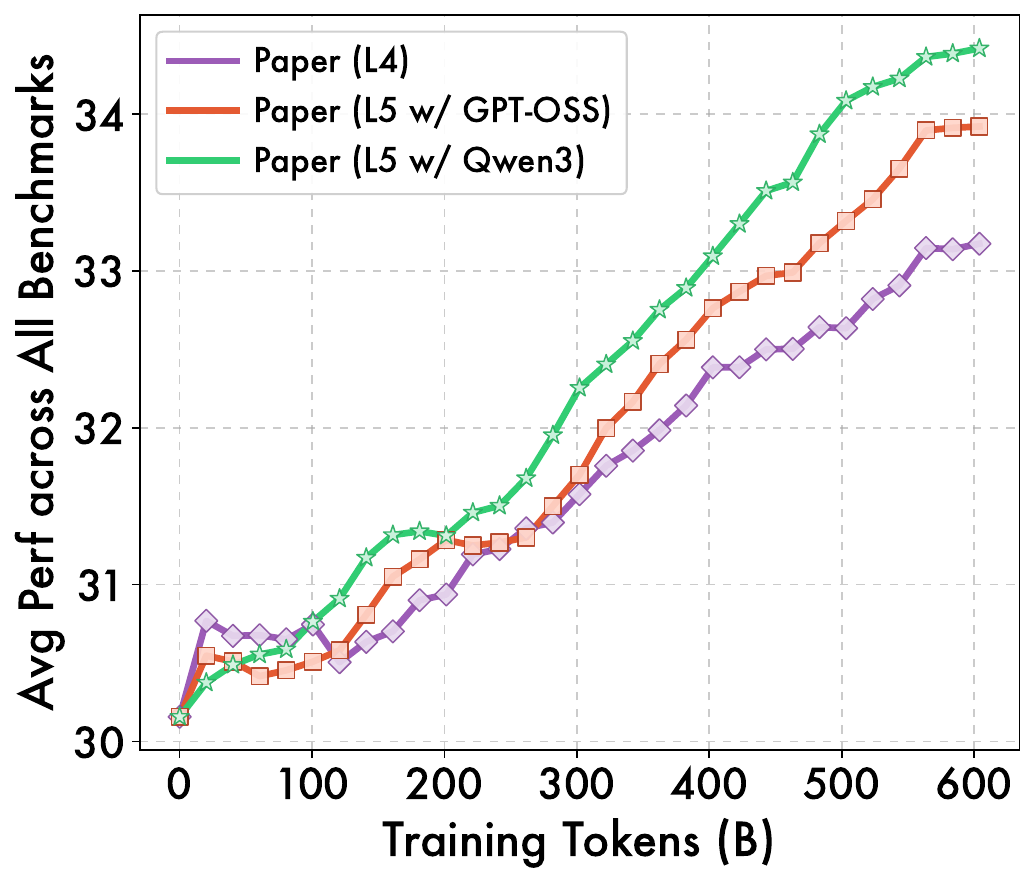}
        \caption{Processing Strategy} 
        \label{fig:proc_strategy}
    \end{subfigure}
    \hfill
    \begin{subfigure}{0.305\textwidth}
        \centering
        \includegraphics[width=\linewidth]{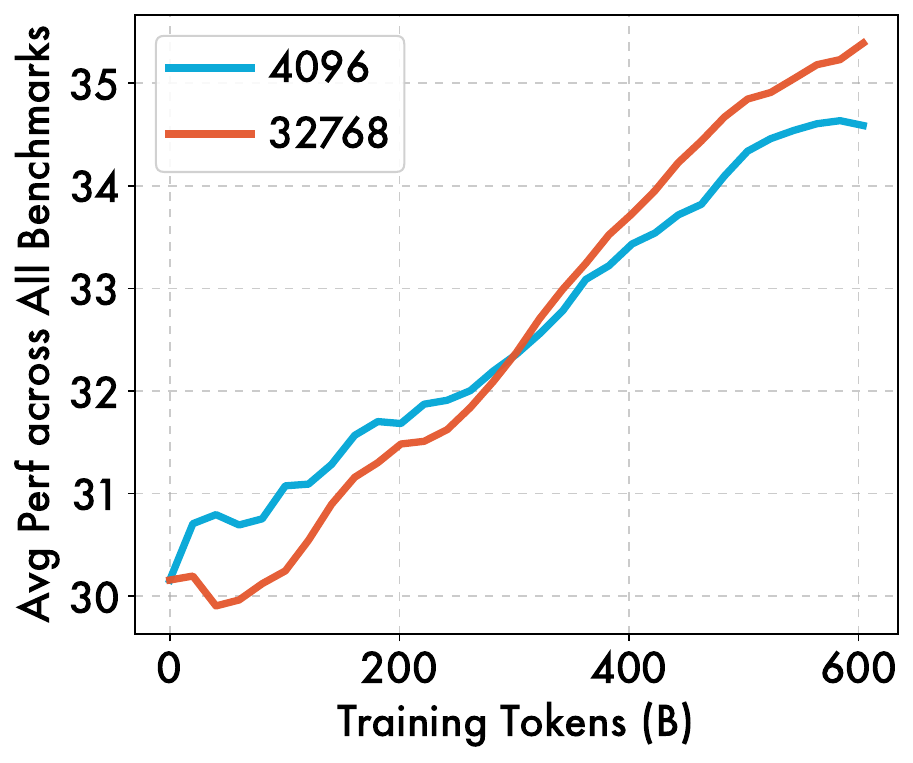}
        \caption{Context Length}
        \label{fig:context_len}
    \end{subfigure}
    \hfill
    \begin{subfigure}{0.305\textwidth}
        \centering
        \includegraphics[width=\linewidth]{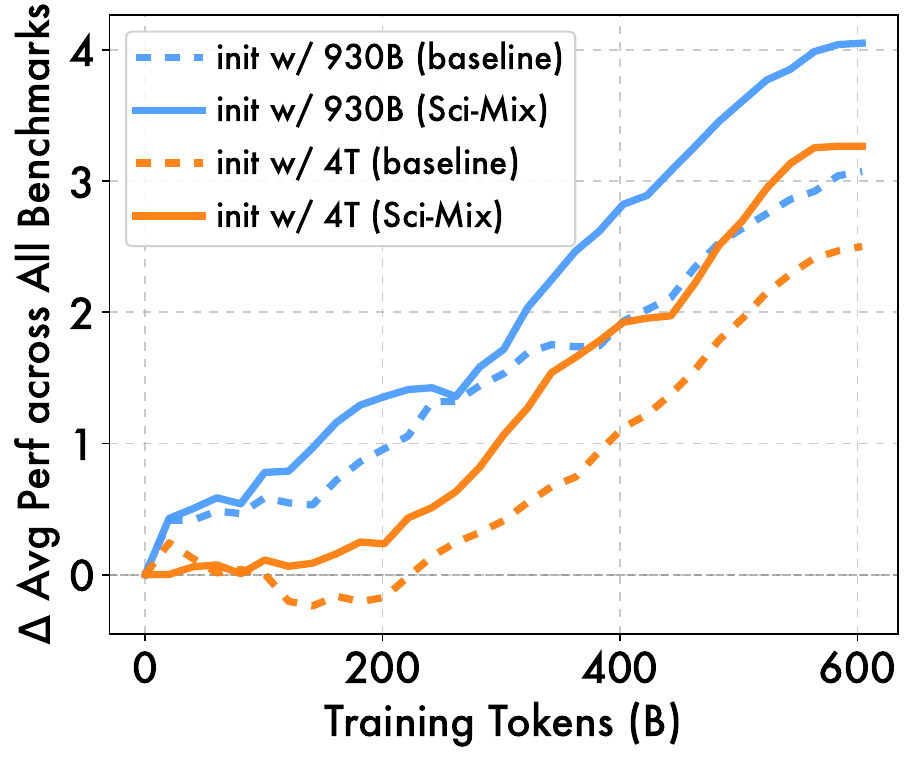}
        \caption{Training Checkpoints}
        \label{fig:train_checkpoints}
    \end{subfigure}

    \caption{(a) Dissection of processing strategies on scientific papers, contrasting content cleaning versus pedagogical augmentation. (b) Comparison of models trained with different context lengths. (c) Performance gains of Sci-Mix over the baseline starting from different training base checkpoints.}
    \label{fig:combined_analysis}
\end{figure*}


To move beyond validation to \textit{optimal training recipes}, we investigate the mechanisms underlying this success. We systematically analyze both \textbf{Data-Centric} and \textbf{Model-Centric} factors through controlled ablations 
on \textit{daVinci-origin-3B} to isolate key drivers and establish evidence-based guidelines.

\subsection{Data-Centric Analysis}
\label{sec:data_factors}

We examine two fundamental dimensions of data preparation: the Composition Strategy to optimize data mixtures, and the Processing Strategy to maximize content learnability.


\subsubsection{Composition Strategy}
\paragraph{Scientific Content Ratio}



We evaluate scientific ratios from 15\% to 100\% (1:1 books-to-papers) and find that aggregated benchmarks follow an inverted-U pattern peaking at 50\% (Fig.~\ref{fig:ratio_all_bench}). Pure scientific training lags behind this balanced mixture, suggesting that \textit{general-purpose performance requires balancing domain focus with broad capabilities}. Specifically, ratios below 30\% offer insufficient domain exposure, while excessive scientific data degrades general reasoning.

Conversely, aligned benchmark performance increases monotonically with scientific ratio (Fig.~\ref{fig:ratio_scipedia_eval}). This divergence shows that \textit{optimal composition is goal-dependent}: balanced mixes suit generalists, while specialized applications favor higher proportions. Thus, "saturation" observed on standard metrics may stem from target mismatch rather than purely from inherent data limits.

\paragraph{Book-Paper Balance}


Beyond the overall ratio, the internal composition between books and papers is also critical. Books provide systematic foundational knowledge with pedagogical structure, while papers present cutting-edge research with technical depth. Testing five book:paper ratios (100:0--0:100) at a fixed 50\% scientific content shows stable performance across mixtures but degrades at extremes (Fig.~\ref{fig:internal_ratio}). This suggests that \textit{books and papers provide complementary value}, and the model's relative insensitivity to precise proportions for practical flexibility based on data availability. Consequently, we adopt a 1:2 ratio, which reflects the composition of our acquired data pool.

\subsubsection{Processing Strategy}



To isolate the specific contribution of L5 (Cognitive Completion), we compare L4 (Generative Refinement) papers against their L5 counterparts on identical subsets. Additionally, we employ \texttt{GPT-OSS-120B} and \texttt{Qwen3-235B} as teacher models to assess the impact of generator quality (Fig.~\ref{fig:proc_strategy}).

Both L5 variants surpass the L4 refinement baseline (\texttt{OSS-120B}: +0.75, \texttt{Qwen3-235B}: +1.27), confirming that \textit{cognitive completion adds distinct value beyond generative refinement}. Furthermore, \texttt{Qwen3-235B} yields an additional +0.52 gain over \texttt{OSS-120B}, demonstrating that \textit{teacher model quality is a critical determinant of cognitive completion effectiveness}.


\subsection{Model-Centric Analysis}
\label{sec:model_factors}

Data learnability is not intrinsic; it is also determined by the learner. Beyond model scale (discussed in Sec.~\ref{sec:main_results}), we investigate two additional properties that affect learning: 

\paragraph{Context Length Requirements}


Since scientific reasoning involves long-range dependencies, we compare a standard 4K context window (RoPE base = 10,000) against an extended 32K (RoPE base = 1,000,000), finding that the 32K ultimately leads by \textbf{+0.80} points (Fig.~\ref{fig:context_len}). Learning dynamics reveal an initial adaptation phase: while the 4K model leads early, the 32K version progressively pulls ahead, implying that \textit{extended context yields superior long-term performance but requires adaptation time}. Practitioners should thus evaluate over sufficient training durations, as long-context advantages emerge gradually.

\paragraph{Training Stage Consistency} 
We investigate whether the benefits of scientific data depend on model maturity by comparing early-stage (930B tokens) vs. late-stage (4T tokens) checkpoints, both continuing training for 600B tokens with Baseline and Sci-Mix configurations (Fig.~\ref{fig:train_checkpoints}).

Both stages exhibit robust improvements over their respective baselines (Early: +0.98, Late: +0.76). This consistency yields two insights. First, the persistent gain at the late stage confirms that \textit{scientific data remains effective even for mature models}. Second, the comparable magnitude of these gains implies that \textit{early checkpoints serve as reliable proxies for data evaluation}, enabling corpus assessment at a fraction of the compute cost.

\section{Related Work}

\paragraph{Domain-Specialized Pre-training Data for Science.}
A central challenge in large-scale pre-training lies in the scarcity of high-quality, domain-specific corpora beyond general-purpose web data. Open-domain resources such as C4~\citep{raffel2020exploring}, RefinedWeb~\citep{penedo2023refinedweb}, Dolma~\citep{soldaini2024dolma}, and FineWeb~\citep{penedo2024fineweb} have established scalable, high-quality web curation pipelines and inspired controlled data-mix ablations~\citep{li2024datacomp}. Building upon these foundations, the mathematics domain has become the clearest exemplar of specialized pre-training. Notable corpora such as OpenWebMath~\citep{paster2023openwebmath}, MathPile~\citep{wang2024mathpile}, InfiMM-WebMath-40B~\citep{han2024infimm}, MegaMath~\citep{zhou2025megamath}, and code-augmented MathCoder2~\citep{lu2024mathcoder2} demonstrate how targeted curation and continued pre-training can significantly enhance reasoning capability. Meanwhile, instruction/SFT resources such as OpenMathInstruct-2~\citep{toshniwal2024openmathinstruct} and Skywork-Math~\citep{zeng2024skywork} extend this paradigm into supervised domains but focus primarily on post-training rather than foundational corpus construction. Beyond mathematics, recent scientific efforts like MegaScience~\citep{fan2025megascience} explore science reasoning and question generation from textbooks and curated Q/A, and evaluate with benchmarks such as GPQA~\citep{rein2024gpqa} and MMLU~\citep{hendryckstest2021}. However, across physics, chemistry, biology, and broader STEM disciplines, there remains a pronounced gap: the community still lacks an open, richly parsed, multi-discipline corpus of \textbf{high-density, cognitively demanding, research-grade scientific materials}---texts that encode complex conceptual reasoning and are suitable for use in the early stages of model development (pre-training and mid-training). Such corpora are essential for teaching models deep scientific abstraction and reasoning patterns, yet remain strikingly underexplored.

\paragraph{Pre-training Data Processing and Transformation.}
Modern pre-training pipelines typically progress through a hierarchy: from raw source parsing (HTML/PDF extraction), to rule-based filtering and deduplication (length, charset heuristics, MinHash/n-gram), to model-based filtering and selection (perplexity, learned raters, or influence-guided resampling), and finally to LLM-driven transformation (reformatting, rephrasing, or repairing) and reasoning-aware augmentation. The early stages are well documented in large open corpora such as FineWeb~\citep{penedo2024fineweb} and Dolma~\citep{soldaini2024dolma}, as well as in classical deduplication methods~\citep{broder1997resemblance}. More recent studies advance the upper stages of this hierarchy: ProX treats refinement as “programming every example” with small models executing fine-grained edits~\citep{zhou2024programming}; RefineX~\citep{bi2025refinex} formalizes expert-guided edit programs for scalable corpus surgery; Nemotron-CC~\citep{su2024nemotron} integrates classifier ensembles with synthetic rephrasing to balance scale and quality; WRAP~\citep{maini2024rephrasing} rephrases web text into QA/Wikipedia-like forms for efficiency; and Generative Data Refinement~\citep{jiang2025generative} leverages LLMs for structured rewriting, detoxification, and anonymization. Parallel research on workflow automation~\citep{li2024autodcworkflow} and LLM-based data cleaning~\citep{zhang2025data} underscores growing interest in model-assisted curation. 
Despite this progress, existing pipelines remain largely web-oriented and seldom address the unique challenges of \textbf{scientific books and research papers}, which contain highly technical, symbol-rich, and conceptually abstract expressions. To date, no large-scale system has combined (i) comprehensive collection of scientific literature, (ii) multi-stage LLM-based filtering and semantic cleaning, and (iii) pedagogical rewriting that transforms dense expert-level prose into content more interpretable for language models. Our work closes this gap by operationalizing this full hierarchy on scientific texts and systematically studying how stage, ratio, and model choices affect downstream learning.

\section{Conclusion}

This work addresses a fundamental gap in foundation model research: the absence of systematic principles for data processing. We introduce \textbf{Data Darwinism}, a hierarchical framework (L0--L9) that organizes data transformations along three dimensions---selection to generation, preservation to transformation, human-centric to machine-driven---and conceptualizes data quality as co-evolving with model capabilities rather than a static property.

By implementing levels L0--L5 of this framework on scientific literature, we construct \texttt{Darwin-Science}, a 900B-token corpus.
Our investigation reveals that raw scientific data suffers a severe learnability gap: despite high information density, unprocessed content provides negligible training value. 
Systematic hierarchy ascension is essential: while basic filtering (L0–L3) yields negligible results, advancing through L4 (generative refinement) to L5 (cognitive completion) achieves a cumulative gain of +1.36 points by purifying content, making implicit reasoning explicit and adding pedagogical scaffolding.

Through controlled experiments with contamination-free \textit{daVinci-origin} baselines and 600B continued pre-training tokens, we demonstrate that \texttt{Darwin-Science} outperforms competitive mixtures by +2.12 (3B) and +2.95 (7B) points on general benchmarks, amplifying to +5.60 and +8.40 on domain-aligned evaluation. Performance sustains without saturation; larger models extract disproportionate value; and domain-matched assessment reveals 3$\times$ stronger signals than standard benchmarks.

Our findings establish evidence-based guidelines: 50\% scientific content optimizes domain-general balance; teacher model quality determines cognitive completion effectiveness; extended context provides measurable advantages; and hierarchy-driven processing unlocks latent value that raw data cannot deliver. By releasing \texttt{Darwin-Science}, \textit{daVinci-origin} models, and \texttt{Darwin-Science-Eval}, we provide both conceptual foundations and practical resources for principled data-model co-evolution.

\paragraph{Limitations and Future Work.} This work focuses on scientific domains and implements L0--L5; higher levels (L6--L9) involving multi-step reasoning synthesis, personalized curriculum generation, and world simulation remain unexplored. Our experiments use specific teacher models and training configurations; broader ablations across architectures, scales, and domains would strengthen generalizability. The learnability gap phenomenon warrants deeper investigation into what makes content machine-learnable versus human-readable.

Data Darwinism represents a first step toward systematic data science for AI. As models continue advancing, the framework's co-evolutionary perspective---where better models enable better data, which trains better models---offers a principled path for sustained progress. We envision future work extending this hierarchy to multimodal domains, formalizing learnability metrics, and developing automated systems that navigate the full L0--L9 spectrum to unlock value from humanity's accumulated knowledge.

\bibliographystyle{acl_natbib}
\bibliography{bib}
\clearpage
\appendix

\clearpage
\appendix

\section{Classification}

\subsection{Discipline Classification Mapping Rules}
\label{app:fdc-mapping}
The Dewey Decimal Classification (DDC) is a widely adopted library classification system that systematically organizes knowledge through decimal numerical codes. It employs a hierarchical structure where the hundreds digit represents the main class, the tens and ones digits subdivide into subclasses, and digits after the decimal point provide finer granularity, theoretically supporting hierarchical subdivision to arbitrary depth. While this natural hierarchical structure facilitates discipline classification, the classification granularity is overly fine-grained, and portions of the system originate from historical periods that do not adequately reflect contemporary disciplinary development and evolution. Therefore, we merged and remapped the numerical codes from FDC labels to align them with disciplines suitable for current research needs.

\begin{longtable}{lll}
\toprule
\textbf{Higher Level Category } & \textbf{Code Range} & \textbf{Category} \\
\midrule
\endfirsthead
\multicolumn{3}{c}{\tablename\ \thetable{} -- continued} \\
\toprule
\textbf{Higher Level} & \textbf{Code Range} & \textbf{Category} \\
\midrule
\endhead
\bottomrule
\endfoot
\bottomrule
\endlastfoot

\rowcolor{blue!8}
\textbf{computer science} & 000-009 & computer\_science \\

\rowcolor{white}
\textbf{engineer} & 355-359 & military\_science \\
\rowcolor{white}
 & 600-610, 620-621, 626, 629 & engineering \\
\rowcolor{white}
 & 622 & engineering\_mining \\
\rowcolor{white}
 & 623 & engineering\_maritime \\
\rowcolor{white}
 & 624 & engineering\_civil \\
\rowcolor{white}
 & 625 & engineering\_railway \\
\rowcolor{white}
 & 627 & engineering\_water \\
\rowcolor{white}
 & 628 & engineering\_environment \\
\rowcolor{white}
 & 630-631, 632-635, 636-639 & agriculture \\
\rowcolor{white}
 & 660-669 & engineering\_chemical \\
\rowcolor{white}
 & 670-689 & manufacturing \\
\rowcolor{white}
 & 690-699 & construction \\

\rowcolor{blue!8}
\textbf{mathematics} & 500-519 & mathematics \\

\rowcolor{white}
\textbf{physics} & 530-539 & physics \\

\rowcolor{blue!8}
\textbf{chemistry} & 540-549 & chemistry \\

\rowcolor{white}
\textbf{biology} & 570-579 & biology \\

\rowcolor{blue!8}
\textbf{medicine} & 610-619 & medicine \\

\rowcolor{white}
\textbf{stem-others} & 520-529 & natural\_sciences\_astronomy \\
\rowcolor{white}
 & 550-559 & natural\_sciences\_earth \\
\rowcolor{white}
 & 560-569 & natural\_sciences\_paleontology \\
\rowcolor{white}
 & 580-589 & natural\_sciences\_botany \\
\rowcolor{white}
 & 590-599 & natural\_sciences\_zoology \\
\rowcolor{white}
 & 910-919 & natural\_sciences\_geography \\

\rowcolor{blue!8}
\textbf{humansocial} & 010-099, 350-354, 640-649, 650-659 & management \\
\rowcolor{blue!8}
 & 100-129, 140-149, 160-199 & philosophy \\
\rowcolor{blue!8}
 & 130-139, 150-159 & psychology \\
\rowcolor{blue!8}
 & 200-299 & religion \\
\rowcolor{blue!8}
 & 300-319, 360-369, 380-399 & sociology \\
\rowcolor{blue!8}
 & 320-329 & political\_science \\
\rowcolor{blue!8}
 & 330-339 & economics \\
\rowcolor{blue!8}
 & 340-349 & law \\
\rowcolor{blue!8}
 & 370-379 & education \\
\rowcolor{blue!8}
 & 400-499 & linguistics \\
\rowcolor{blue!8}
 & 700-709, 750-769 & art\_fine\_arts \\
\rowcolor{blue!8}
 & 710-729 & art\_architecture \\
\rowcolor{blue!8}
 & 730-739 & art\_artifacts \\
\rowcolor{blue!8}
 & 740-749 & art\_design \\
\rowcolor{blue!8}
 & 770-779 & art\_photography \\
\rowcolor{blue!8}
 & 780-789 & art\_music \\
\rowcolor{blue!8}
 & 790-799 & art\_sports \\
\rowcolor{blue!8}
 & 800-899 & literature \\
\rowcolor{blue!8}
 & 900-909, 920-999 & history \\
\end{longtable}

\subsection{Book-Paper classification}
\label{app:book-paper-split}
Given the significant differences in knowledge density between books and papers, we first need to distinguish between these two types of documents to implement targeted processing strategies. We employ the Qwen2.5-7B-Instruct  for this classification task, with the prompt design as follows:

\begin{tcolorbox}[
    colback=blue!5!white,
    colframe=blue!30!gray,
    boxrule=2pt,
    arc=8pt,
    title={\textbf{Book Paper Split Prompt}},
    fonttitle=\large\bfseries
]

\begin{verbatim}
Determine if this document is a scientific academic paper.
    
Note: The following is a sampled portion of a larger document.

Look for:
- Scientific research content with technical depth
- Formal academic writing style
- Dense technical terminology and concepts
- Complex analytical content

Exclude:
- News articles, interviews
- Blog posts, web content
- Documentation, manuals
- Simple explanatory content

Text sample from document:
{text_sample}

Please strictly return the result in the following JSON format,
do not add any other content:
{
    "analysis": "analysis of why this is or isn't an academic 
                 paper with sufficient complexity",
    "is_article": true/false
}
\end{verbatim}

\end{tcolorbox}

\subsection{Darwin-Science Domain Distribution}
\label{app:scipedia-distribution}
As mentioned earlier, we categorize SciPedia by discipline. The detailed domain distribution is provided in Table ~\ref{tab:book_domain_tokens} and Table ~\ref{tab:paper_domain_tokens}.
\begin{table}[htbp]
\centering
\small
\caption{Token Distribution by Domain in Book}
\label{tab:book_domain_tokens}
\begin{tabular}{lrr}
\toprule
\textbf{Domain} & \textbf{Tokens (B)} & \textbf{Percentage} \\
\midrule
Computer Science       & 10.52  & 4.18\%  \\
Engineering            & 22.19  & 8.83\%  \\
Human \& Social        & 148.43 & 59.02\% \\
Medicine               & 27.79  & 11.05\% \\
Biology                & 8.44   & 3.36\%  \\
Chemistry              & 7.14   & 2.84\%  \\
Mathematics            & 11.18  & 4.44\%  \\
Physics                & 4.69   & 1.86\%  \\
STEM Others            & 11.12  & 4.42\%  \\
\midrule
\textbf{Total}         & \textbf{251.49} & \textbf{100.00\%} \\
\bottomrule
\end{tabular}
\end{table}

\begin{table}[htbp]
\centering
\small
\caption{Token Distribution by Domain in Paper}
\label{tab:paper_domain_tokens}
\begin{tabular}{lrr}
\toprule
\textbf{Domain} & \textbf{Tokens (B)} & \textbf{Percentage} \\
\midrule
Computer Science       & 49.90  & 7.63\%  \\
Engineering            & 38.03  & 5.82\%  \\
Human \& Social        & 45.35 & 6.93\% \\
Medicine               & 255.05  & 38.87\% \\
Biology                & 58.28 & 8.91\%  \\
Chemistry              & 42.85   & 6.55\%  \\
Mathematics            & 77.29  & 11.81\%  \\
Physics                & 57.49   & 8.79\%  \\
STEM Others            & 30.71  & 4.69\%  \\
\midrule
\textbf{Total}         & \textbf{655} & \textbf{100.00\%} \\
\bottomrule
\end{tabular}
\end{table}

\section{L4 Processing Details}
\label{app:l3_details}
This appendix provides comprehensive supplementary materials for L4 processing, including the empirical analysis that informed our cleaning protocol design, complete prompt specifications, representative cleaning examples, and evaluation protocols.

\subsection{L4 Processing Prompt}
\label{app:l3_prompt}

To ensure the L4 refinement rules were grounded in the actual quality characteristics of our scientific corpus, we performed an empirical analysis on a random sample of 20 documents, generating 40 detailed assessment reports via \texttt{Gemini 2.5 Pro} and \texttt{Claude Sonnet 4.0}. The recurring quality issues identified in these reports were synthesized into two core operational pillars: Deletion (removing extraneous, non-educational noise) and Modification (repairing and standardizing structural defects). This section presents the resulting production prompt, which codifies these data-driven insights into explicit processing rules and content protection guidelines designed to purify the text while strictly preserving academic integrity.

\begin{tcolorbox}[
colback=blue!5!white,
colframe=blue!30!gray,
  boxrule=2pt,
  arc=8pt,
  title={\textbf{L4 Processing Prompt}},
  fonttitle=\large\bfseries,
  enhanced jigsaw,
  breakable
]
\begin{TextBlock}
You are an expert document cleaner specialized in identifying and removing unwanted content and correcting OCR errors from various document (mainly academic) chunks.

## Objective:
Clean and standardize OCR text by identifying and removing redundant, erroneous, or unwanted content and correcting obvious OCR errors according to the rules below. Your task is to identify and delete unnecessary content completely, fix technical errors, while preserving all academic value.

## Deletion and Correction Rules:
### Document Structural Deletion
* Remove **table of contents and navigation structures**: Multiple consecutive chapter/section titles listed together without accompanying text content
  - **Preserve content section headings in main text**: such as chapter headings, section titles followed by explanatory text or academic material
* Remove **reference lists completely**: numbered entries with author names, publication titles, and years (e.g., "1. Smith, J. (2020). Title. Journal, 15(3), 123-145.") **[Delete entire list regardless of format]**
* Remove **front matter and back matter**: such as prefaces, acknowledgments, copyright statements, indexes, and other standard book structural elements
  - **Preserve sections with academic value**: such as abstracts, introductions, conclusions that present research background or methodology
* Remove **publication and metadata information**: such as ISBN, publisher information, revision history, version numbers, institutional affiliations, author affiliations, addresses, contact information
* Remove **page headers, page footers, and page numbers**

### Academic Content Deletion
* Remove **pure indexing appendices**: such as glossaries, symbol tables, abbreviation lists, indexes, notations and other purely referential lookup content (entries that only provide definitions without explanations, e.g., "a - alpha coefficient")
  - **Preserve**: appendices with learning value (e.g. mathematical derivations, proofs, technical explanations)
  - **Preserve**: explanatory content that directly supports main text elements (e.g. abbreviation/parameter explanations after tables/formulas/diagrams)
* Remove **image files and placeholders**: such as `<img>` tags, image file paths, image URLs, markdown syntax and image placeholders (e.g. `[Image]`, `[Picture not available]`)
  - **Preserve**: figure/table titles, descriptive text (including content within markdown image formats: ![description](path) → description)
  - **Preserve**: in-text references (e.g., "as shown in Figure 1")

### Invalid and Redundant Content Deletion
* Remove **OCR processing artifacts**: such as garbled text, encoding artifacts, duplicate characters, malformed special characters, OCR messages (`[OCR error]`), file paths, timestamps, version numbers, revision history
* Remove **garbage content**: such as junk information, advertising content, placeholders (e.g. [Insert citation here])
* Remove **duplicate content**: identical paragraphs or sections mainly caused by OCR errors
  - **Exception**: Carefully apply to technical formulas, equations, or specialized notation that may contain subtle but meaningful differences
  - **Exception**: Apply contextual analysis - preserve identical content that serves different semantic purposes or artistic purposes (e.g., poetic refrains, literary repetition)
* Remove **content and navigation markers**: [content missing], [page break], (Continued), and similar placeholder markers
* Remove **URLs and links**: all web addresses, hyperlinks, and link information

### OCR Error Correction
* **Fix text fragmentation**: repair split words, broken sentences, erroneous line breaks and paragraph divisions, missing spaces and punctuation
* **Fix fragmented structured content**: Repair OCR-damaged structured content (e.g. tables, diagrams, formulas) appearing as consecutive lines of isolated words, single characters, or short phrases
  - **Pattern**: Consecutive lines (5+) with 1-3 words/characters each
  - **Action**: Preserve content while indicating structural damage; delete if unrepairable
* **Standardize whitespace and formatting**: clean excessive whitespace, compress blank lines, standardize spacing and indentation
* **Fix character and encoding errors**: correct obvious character errors, spelling issues, and Unicode anomalies
* **Standardize punctuation**: unify quotation marks, dashes, hyphens, and other punctuation
* **Complete truncated words**: only fix obviously incomplete words from clear OCR errors, avoid modifying content at chunk edges
* **Standardize academic formatting**: remove excessive LaTeX commands and unify notation format

## Content Protection Rules:
### Always Preserve Academic and Educational Content
* Preserve **Technical and specialized content**: such as formulas, equations, proofs, symbols, chemical structures, biological sequences and their original format
  - **Preserve exact content**: do not alter variables, coefficients, structures, sequences, or any technical details
* Preserve **In-text references and citations**: such as (Smith, 2020), [15], "see Chapter 2", equation (5), "Figure 2.5", (pp. 3-7)
* Preserve **Table structures**: preserve academic table content, formatting and structural markers (e.g. "|", HTML tags)
  - **Exception**: Does not apply to navigation tables (table of contents, indexes, glossaries) which should be removed
* Preserve **Code blocks and programming examples**: preserve code block markers (```language, ```, etc.) and internal code syntax and structure
* Preserve **Educational content**: such as exercises, questions, answers, solutions, case studies, instructions, user guides
* Preserve **Explanatory content**: such as NOTE boxes, WARNING boxes, tips, author comments, supplementary information, academic footnotes
* Preserve **Chunk boundary content**: incomplete sentences and words at chunk edges due to text segmentation
* Preserve **Literary and humanities content**: including poetry, fiction, drama, creative writing, literary analysis, philosophical texts, and other humanities scholarship with educational value

## Instructions:
- Carefully identify all content matching the deletion rules
- Remove completely any content that should be deleted
- Preserve all valuable academic content by applying protection rules and retaining content that doesn't match deletion rules
- Apply OCR error corrections to fix obvious technical problems
- Ensure text flows naturally after corrections and deletions
- If the entire chunk should be deleted, leave the output tags completely empty
- **Important**: The content inside the <CLEANED_TEXT> tags must be exactly the text after deletion, with no explanations, comments, or additional text inside the tags

## Input:
OCR document chunk:
[CHUNK]

## Output Format:
<CLEANED_TEXT>
[Place the cleaned content here, or leave completely empty if everything should be deleted]
</CLEANED_TEXT>
\end{TextBlock}
\end{tcolorbox}

\subsection{Evaluation and Model Selection}
\label{app:l3_evaluation_and_model_section}
To ensure effective content cleaning and support iterative prompt refinement, we developed a comprehensive evaluation framework. Given that content cleaning operates through explicit deletion and correction rules, evaluation must assess both rule execution accuracy (whether rules are correctly applied) and rule completeness (whether the rule set covers all necessary cases). We adopted a hybrid strategy combining human inspection for identifying improvement opportunities with LLM-based evaluation for systematic quality assessment and quantitative comparison across different prompts and cleaning models.

\paragraph{Evaluation Dataset} We randomly sampled 20 documents from our corpus to serve as representative 
evaluation cases, ensuring diversity across scientific domains and document types (books vs. papers). 

\paragraph{Human Evaluation}
Human evaluators reviewed entire documents, with particular attention to high-risk sections—document beginnings and endings where table of contents, reference lists, and structural artifacts typically appear. Evaluators identified execution failures and uncovered problematic content types not addressed by current rules, providing qualitative feedback that directly informed prompt iterations.

\paragraph{LLM-based Evaluation}

To enable a systematic comparison of L4 refinement quality across different models and prompts, we employ an LLM-as-a-judge methodology using \texttt{Claude-Sonnet-4.0}~\citep{anthropic2025claude4systemcard} and \texttt{Gemini-2.5-Pro}~\citep{comanici2025gemini}. Our evaluation corpus consists of three groups of three consecutive chunks sampled from 20 representative documents to ensure local coherence.

The evaluation is guided by the following prompt that instructs judges to perform a rule-by-rule analysis of execution accuracy, identify coverage gaps, and document concrete examples of failures. By generating a structured output that includes both a quantitative quality score and prioritized recommendations, this approach enables rigorous performance comparisons while simultaneously gathering the actionable insights necessary for iterative rule refinement.



\label{app:l3_eval_prompt}
\begin{tcolorbox}[
colback=blue!5!white,
colframe=blue!30!gray,
  boxrule=2pt,
  arc=8pt,
  title={\textbf{L4 Evaluation Prompt}},
  fonttitle=\large\bfseries,
  enhanced jigsaw,
  breakable
]
\begin{TextBlock}
# Data Cleaning Quality Evaluation Prompt

You are an expert in data preprocessing and text cleaning quality assessment. Your task is to evaluate text data cleaning quality by analyzing rule execution accuracy and rule completeness. Focus specifically on the deletion and OCR error correction phases of cleaning - identifying what was done incorrectly and what rules need improvement.

## Evaluation Focus
1. **Rule Execution Accuracy**: Check if cleaning rules were correctly applied to identify and remove unwanted content, and if OCR correction rules were properly applied to fix technical errors
2. **Rule Completeness**: Assess if the cleaning rules cover all necessary cases and are clearly defined

Note: This evaluation focuses on deletion and OCR error correction phases. Advanced text modification (restructuring, rewriting, semantic improvements) happens in a separate step and should not be included in the scoring, but suggestions can be provided.

## Input:
Cleaning rules:
[CLEAN_RULES]

Text samples before and after cleaning:
[EVALUATION_INPUT]

## Output Format
```markdown
# Data Cleaning Quality Evaluation Report

## 1. Rule Execution Accuracy Analysis (By Rule)

*Evaluate each cleaning rule individually. Every rule must be assessed, even if it was executed perfectly.*

### Rule 1: [Rule Name/Description]
**Execution Quality:** [Excellent/Good/Fair/Poor]
**Missed Deletions/Corrections:** [Number] instances (0 if none)
**Incorrect Deletions/Corrections:** [Number] instances (0 if none)

**Examples (if any issues):**
```
Chunk ID: [specify the chunk ID from the evaluation input]

Before cleaning: [copy exactly from the original text provided]

After cleaning: [copy exactly from the cleaned text provided - must be ACTUAL result, not what you think it should be]

Problem: [highlight specific issues]
Explanation: [why this is problematic according to the rule]
```

*If no issues: "No issues found - this rule was executed correctly throughout the text."*

### Rule 2: [Rule Name/Description]
**Execution Quality:** [Excellent/Good/Fair/Poor]
**Missed Deletions/Corrections:** [Number] instances (0 if none)
**Incorrect Deletions/Corrections:** [Number] instances (0 if none)

*[Continue for EVERY cleaning rule provided - do not skip any rules]*

### Overall Execution Summary
**Rules with Most Issues:**
1. [Rule name] - [X missed deletions/corrections, X incorrect deletions/corrections]
2. [Rule name] - [X missed deletions/corrections, X incorrect deletions/corrections]
3. [Rule name] - [X missed deletions/corrections, X incorrect deletions/corrections]

## 2. Rule Completeness Analysis

### 2.1 Missing Rules (New cleaning needs discovered)
**Content type found:** [Describe unwanted content or OCR errors that current rules don't address]
**Suggested new rule:** [Specific rule to handle this content/error]

**Example:**
```
Chunk ID: [specify the chunk ID from the evaluation input]

Problematic content found: [show the unwanted content or OCR error in text]

How it should be cleaned: [show desired result]
```

### 2.2 Existing Rules Needing Improvement
**Rule name:** [Specific rule that needs changes]
**Problem:** [What's wrong - ambiguity, inaccuracy, or other issues]
**Suggested improvement:** [How to fix the rule - modification or clarification]

**Example:**
```
Chunk ID: [specify the chunk ID from the evaluation input]

Current rule causes: [problematic cleaning result or inconsistent application]

After improvement should be: [improved result]
```

## 3. Additional Observations

### Advanced Modification Suggestions (Not scored)
*Note: These are suggestions for advanced modification phase (restructuring, rewriting, semantic improvements) and do not affect the current cleaning quality score.*

[Any suggestions for advanced text modification/restructuring improvements that should be handled in the next phase]

## 4. Evaluation Summary

### Overall Cleaning Quality
[Excellent/Good/Fair/Poor] - [Brief explanation of assessment reasoning based on deletion and OCR correction accuracy]

### Issues and Recommendations by Priority

**High Priority:** [Problems that significantly impact deletion or OCR correction accuracy]
- Issues: [specific problems]
- Recommendations: [concrete solutions]

**Medium Priority:** [Problems that affect cleaning consistency but not core quality]  
- Issues: [specific problems]
- Recommendations: [concrete solutions]

**Low Priority:** [Minor cleaning optimization opportunities]
- Issues: [specific problems] 
- Recommendations: [concrete solutions]
```

## Important Note
Provide honest assessment based on actual observations. Focus on whether content that should be deleted was correctly identified and removed, and whether OCR errors were properly corrected. If the cleaning quality is excellent with minimal issues, report that truthfully. Don't artificially identify problems - accurate evaluation is more valuable than finding issues where none exist.

Note that not every section in the report needs to be filled. If there are no issues in a particular category, you can leave that section empty or state "No issues found in this category."
\end{TextBlock}
\end{tcolorbox}

\paragraph{Model Selection}
We compared multiple language models for Content Cleaning cleaning using identical prompts and evaluation protocols: Qwen2.5 series (7B, 32B, 72B-Instruct)~\citep{qwen2.5}, \texttt{Llama3.3-70B-Instruct}~\citep{meta2024llama33}, Qwen3 series (8B, 14B, 32B, 235B, both thinking and non-thinking variants)~\citep{qwen3technicalreport}, and \texttt{GPT-OSS-120B}~\citep{openai2025gptoss120bgptoss20bmodel}.

Our evaluation results show that Qwen3 series substantially outperformed Qwen2.5 and Llama3.3. Within Qwen3, thinking mode achieved higher accuracy but reduced throughput several-fold. Among Qwen3 models, 8B underperformed while 14B, 32B, and 235B showed comparable quality. \texttt{GPT-OSS-120B} demonstrated competitive cleaning accuracy while offering superior processing efficiency, making it our choice for production deployment.

\subsection{Implementation Details}
\label{app:l3_implementation_details}

\paragraph{Quality Control}
\label{app:l3_quality_control}
Not all model outputs are correct. Common failure modes include malformed output formats that prevent proper extraction of cleaned text, infinite repetition until reaching output length limits, and other processing errors. When such failures occur, we retain the original chunk unchanged. A document is considered successfully processed if at least 95\% of its chunks are correctly cleaned; otherwise, it is marked as failed and queued for reprocessing.

\paragraph{Distributed Processing System}
\label{app:l3_distributed} 
Processing pretraining-scale data requires flexible, scalable, 
and robust infrastructure. We adopt a producer-consumer 
architecture where a Redis~\citep{redis} server acts as the 
task queue and GPU servers function as workers running vLLM 
servers that continuously fetch and process tasks.

This design addresses several critical challenges:
\begin{itemize} 
    \item \textbf{Dynamic resource allocation}: The availability of GPU nodes in our cluster varies over time. Our design allows seamless addition or removal of worker nodes without interrupting the overall pipeline. 
    \item \textbf{Orphan task management}: GPU servers may crash or be shut down unexpectedly, leaving tasks incomplete. We implement a heartbeat mechanism to monitor worker health, periodically detecting dead workers and reclaiming their orphaned tasks for reassignment. 
    \item \textbf{Automatic recovery}: vLLM servers running on GPU workers may crash. Our system automatically detects failures and restarts crashed servers to maintain processing continuity.
    \item \textbf{Task retry mechanism}: Tasks that fail due to quality control issues or other errors are automatically re-queued for processing. Tasks exceeding a maximum retry threshold are marked as permanently failed.
    \item \textbf{Priority queuing}: The system supports priority-based task scheduling, allowing high-priority tasks to bypass the standard queue when necessary. 
\end{itemize} 

This architecture enables efficient processing of our large-scale scientific corpus while maintaining robustness against the inevitable failures that occur in distributed systems operating over extended periods.

\subsection{L4 Processing Examples}
\label{app:l3_examples}


This section presents representative before-and-after examples demonstrating L4 cleaning effects on real scientific documents across varying quality levels. Through side-by-side comparisons, we illustrate how L4 processing successfully removes front matter and structural artifacts, standardizes mathematical notation, corrects formatting inconsistencies, recovers text from severe OCR corruption, and preserves all academically valuable content. The examples span from well-formatted thesis documents to heavily damaged scanned texts, showcasing L4's capability to handle diverse quality scenarios common in scientific corpora. Each example highlights specific aspects of the cleaning pipeline, showing the practical impact—and limitations—of our deletion and modification operations on document quality under different degradation conditions.

\vspace{8pt}
\noindent\textbf{Example 1: PhD Thesis Front Matter Cleaning}~
\smallskip

\noindent\emph{This example demonstrates L4 processing on a mathematics PhD thesis, showcasing the removal of typical academic front matter (title page, acknowledgments, table of contents) while preserving research content (abstracts, keywords) and standardizing mathematical notation.}

\begin{CompareBox}[Example 1 (Text Before L4 Processing)]{red}
\small
Towards an ($\infty$,2)-category of homotopy coherent monads in an $\infty$-cosmos\\\\

THÈSE N° 7748 (2017)\\
PRÉSENTÉE LE 22 SEPTEMBRE 2017\\
À LA FACULTÉ DES SCIENCES DE LA VIE\\
LABORATOIRE POUR LA TOPOLOGIE ET LES NEUROSCIENCES\\
PROGRAMME DOCTORAL EN MATHÉMATIQUES\\

ÉCOLE POLYTECHNIQUE FÉDÉRALE DE LAUSANNE\\

POUR L'OBTENTION DU GRADE DE DOCTEUR ÈS SCIENCES\\

PAR\\

Dimitri ZAGANIDIS\\

acceptée sur proposition du jury:\\
Prof. M. Troyanov, président du jury\\
Prof. K. Hess Bellwald, directrice de thèse\\
Prof. D. Verity, rapporteur\\
Prof. E. Riehl, rapporteuse\\
Prof. Z. Patakfalvi, rapporteur\\
I would like to warmly thank my advisor Prof. Kathryn Hess Bellwald for her constant support and encouragement. Thank you for your interest, confidence, optimism and enthusiasm! Thank you for welcoming me in your research group, which provided me with the best work environment I could dream of. You have been a source of inspiration from the beginning.\\

I would like to thank all the past and present members of Kathryn’s group, Varvara Karpova, Marc Stephan, Kay Werndli, Martina Rovelli, Rachel Jeitziner, Justin Young, Gavin Seal, Magdalena Kedziorek, Gard Spreemann, Martina Scolamiero, Jean Verette and Lyne Moser for their friendliness, help and great discussions.\\

A special thank to Jérôme Scherer, who has given me the opportunity to teach geometry to the high potential teenagers of the Euler program. It has been an interesting and challenging experience!\\

I want to warmly thank Prof. Emily Riehl and Prof. Dominic Verity. First of all, I am indebted to them mathematically speaking. The motivation for this thesis came from two different sources. Firstly, an article of Ross Street [50] that we studied in the Kan extension seminar, which was organized by Emily Riehl in 2014. Secondly, from the series of articles by Emily Riehl and Dominic Verity [43, 45, 44]. Dominic Verity is the father of weak complicial sets, the idea to use some of them as models of $(\infty, 2)$-categories came from a lecture by Emily Riehl at the “Higher Structure” conference at MATRIX, Australia, in June 2016. This wonderful conference was organized by Marcy Robertson and Philip Hackney, and I take the opportunity to thank them for organizing such a great event. Finally, I am grateful that Emily Riehl and Dominic Verity accepted to be members of my thesis jury.\\

I also want to thank the other jury members Prof. Zsolt Patakfalvi and Prof. Marc Troyanov for their precious time.\\

I dedicate this thesis to my family, who has always been supportive and encouraging, and in particular to my wife Cynthia, which provides me with so much love and happiness.\\
Abstract\\

This thesis is part of a program initiated by Riehl and Verity to study the category theory of \((\infty, 1)\)-categories in a model-independent way. They showed that most models of \((\infty, 1)\)-categories form an \(\infty\)-cosmos \(\mathcal{K}\), which is essentially a category enriched in quasi-categories with some additional structure reminiscent of a category of fibrant objects. Riehl and Verity showed that it is possible to formulate the category theory of \((\infty, 1)\)-categories directly with \(\infty\)-cosmos axioms. This should also help organize the category theory of \((\infty, 1)\)-categories with structure.\\

Given an \(\infty\)-cosmos \(\mathcal{K}\), we build via a nerve construction a stratified simplicial set \(N_{\text{Mnd}}(\mathcal{K})\) whose objects are homotopy coherent monads in \(\mathcal{K}\). If two \(\infty\)-cosmoi are weakly equivalent, their respective stratified simplicial sets of homotopy coherent monads are also equivalent. This generalizes a construction of Street for 2-categories. We also provide an \((\infty, 2)\)-category \(\text{Adj}_r(\mathcal{K})\) whose objects are homotopy coherent adjunctions in \(\mathcal{K}\), that we use to classify the 1-simplices of \(N_{\text{Mnd}}(\mathcal{K})\) up to homotopy.\\

Key words: higher category, \(\infty\)-cosmos, \((\infty, 2)\)-category, \((\infty, 1)\)-category, homotopy coherent monad, model category\\
Résumé\\

Cette thèse s’inscrit dans un programme initié par Riehl et Verity pour étudier la théorie des ($\infty$, 1)-catégories d’une façon qui ne dépend pas du modèle choisi. Ils ont montré que la plupart des modèles de ($\infty$, 1)-catégories forme un $\infty$-cosmos, c’est-à-dire essentiellement une catégorie enrichie sur les quasi-catégories, munie de plus d’une structure rappelant celle d’une catégorie d’objets fibrants. Riehl et Verity ont montré qu’il est possible de formuler la théorie des catégories satisfaite par les ($\infty$, 1)-catégories directement à partir des axiomes d’$\infty$-cosmos. Ceci devrait également aider à organiser la théorie des ($\infty$, 1)-catégories munies d’une structure.\\

Étant donné un $\infty$-cosmos $\mathcal{K}$, nous construisons, grâce à une construction de nerf, un ensemble simplicial stratifié $\mathcal{N}_{\text{Mnd}}(\mathcal{K})$ dont les objets sont les monades homotopiquement cohérentes dans $\mathcal{K}$. Si deux $\infty$-cosmoi sont faiblement équivalents, leurs ensembles simpliciaux stratifiés des monades homotopiquement cohérentes respectifs sont également équivalents. Ceci généralise une construction de Street pour les 2-catégories. Nous fournissons également une ($\infty$, 2)-catégorie $\text{Adj}_r(\mathcal{K})$ dont les objets sont les adjonctions homotopiquement cohérentes dans $\mathcal{K}$ et que nous utilisons pour classifier les 1-simplexes de $\mathcal{N}_{\text{Mnd}}(\mathcal{K})$ à homotopie près.\\

Mots clefs: catégorie d’ordre supérieur, $\infty$-cosmos, ($\infty$, 2)-catégorie, ($\infty$, 1)-catégorie, monades homotopiquement cohérentes, catégorie de modèles\\
Contents\\

1 Introduction ........................................... 1\\
   1.1 Historical motivations .......................... 1\\
       1.1.1 Monads and adjunctions in classical category theory\\ .... 1\\
       1.1.2 Higher category theory ..................... 2\\
       1.1.3 $\infty$-Cosmoi ............................ 5\\
   1.2 Main contributions and organization of the thesis\\ .......... 6\\

Notations and Terminology ................................ 9\\

2 Background Material .................................. 11\\
   2.1 Enriched categories ........................... 11\\
       2.1.1 Simplicial categories ...................... 11\\
       2.1.2 2-Categories ........................... 16\\
       2.1.3 Weighted limits and enriched right Kan extensions ..... 22\\
   2.2 Higher categories ............................ 26\\
       2.2.1 Quasi-categories ........................ 26\\
       2.2.2 Weak complicial sets ...................... 31\\
   2.3 The universal 2-category containing an adjunction ......... 36\\
       2.3.1 The 2-categorical model .................... 36\\
       2.3.2 The simplicial model ...................... 38\\
       2.3.3 The isomorphism $\mathcal{Adj} \cong \mathcal{Adj}$\\ ................. 41\\
       2.3.4 The Eilenberg-Moore object of algebras as a weighted limit 46\\
   2.4 $\infty$-Cosmoi and their homotopy 2-category .............. 47\\
       2.4.1 Homotopy coherent monads and adjunctions ........ 51\\
       2.4.2 Absolute left liftings and left exact transformations ..... 52\\
       2.4.3 Monadicity theorem ...................... 54\\
   2.5 The homotopy coherent nerve ..................... 58\\
3 The 2-category $\text{Adj}^S_{\text{hc}}[n]$ 65\\

3.1 The 2-category $\text{Adj}[n]$ .......................... 67\\
3.1.1 The 2-categorical model .................... 67\\
3.1.2 The simplicial model ...................... 70\\

3.2 Description of $\text{Adj}^S_{\text{hc}}[n]$ ........................ 73\\
3.2.1 The simplicial and 2-categorical models ........... 73\\
3.2.2 Non-degenerate morphisms of $\text{Adj}^S_{\text{hc}}[n]$ ........... 77\\
3.2.3 Atomic morphisms of $\text{Adj}^S_{\text{hc}}[n]$ ................ 79\\
3.2.4 Convenient 2-subcategories of $\mathcal{C}\Delta[n]^{\text{co}}$ ............ 87\\

4 The lifting theorem 91\\

4.1 Right fillability and parent-child relation ............... 92\\
4.1.1 Right fillable morphism and its distinguished face ...... 92\\
4.1.2 Parent-child relation ...................... 95\\
4.1.3 Right parental relative subcomputads ............ 100\\

4.2 Decomposition as an $L$-cell complex ................. 102\\
4.2.1 A single pushout against a morphism in $L$ .......... 102\\
4.2.2 Transfinite composition .................... 106\\

4.3 Proof of Theorem A .......................... 109\\
4.3.1 Lifting against a morphism of $L_2$ ............... 109\\
4.3.2 The proof ............................ 113\\

5 Homotopy coherent diagrams 115\\

5.1 Universal property of $\text{Adj}^S_{\text{hc}}[n]$ and of $\text{Mnd}^S_{\text{hc}}[n]$ .......... 116\\

5.2 Minimal data defining a ho. coh. diag. of adjunctions ....... 118\\
5.3 Induced ho. coh. diag. of free-forgetful adjunctions ......... 123\\
5.3.1 A projective cell-complex ................... 124\\
5.3.2 Identifying the domain of the right adjoints ......... 129\\

6 Towards an $(\infty, 2)$-category of ho. coh. monads 133\\

6.1 Several stratified nerve constructions ................. 134\\
6.2 Discussion about Conjecture B .................... 137\\
6.3 Proof of Theorem C .......................... 146\\
6.4 Classification results .......................... 149\\
6.4.1 Description of $h(\text{Adj}_r(K)|_1)$ .................. 150\\
6.4.2 A reflective subcategory .................... 153\\

A Proofs of background results 167\\

B Presentation by computads 175\\
\end{CompareBox}

\vspace{6pt}

\begin{CompareBox}[Example 1 (Text After L4 Processing)]{green}
\small
Abstract\\

This thesis is part of a program initiated by Riehl and Verity to study the category theory of \((\infty, 1)\)-categories in a model-independent way. They showed that most models of \((\infty, 1)\)-categories form an \(\infty\)-cosmos \(\mathcal{K}\), which is essentially a category enriched in quasi-categories with some additional structure reminiscent of a category of fibrant objects. Riehl and Verity showed that it is possible to formulate the category theory of \((\infty, 1)\)-categories directly with \(\infty\)-cosmos axioms. This should also help organize the category theory of \((\infty, 1)\)-categories with structure.\\

Given an \(\infty\)-cosmos \(\mathcal{K}\), we build via a nerve construction a stratified simplicial set \(N_{\text{Mnd}}(\mathcal{K})\) whose objects are homotopy coherent monads in \(\mathcal{K}\). If two \(\infty\)-cosmoi are weakly equivalent, their respective stratified simplicial sets of homotopy coherent monads are also equivalent. This generalizes a construction of Street for 2-categories. We also provide an \((\infty, 2)\)-category \(\text{Adj}_r(\mathcal{K})\) whose objects are homotopy coherent adjunctions in \(\mathcal{K}\), that we use to classify the 1-simplices of \(N_{\text{Mnd}}(\mathcal{K})\) up to homotopy.\\

Key words: higher category, \(\infty\)-cosmos, \((\infty, 2)\)-category, \((\infty, 1)\)-category, homotopy coherent monad, model category\\

Résumé\\

Cette thèse s’inscrit dans un programme initié par Riehl et Verity pour étudier la théorie des \((\infty,1)\)-catégories d’une façon qui ne dépend pas du modèle choisi. Ils ont montré que la plupart des modèles de \((\infty,1)\)-catégories forment un \(\infty\)-cosmos, c’est‑à‑dire essentiellement une catégorie enrichie sur les quasi‑catégories, munie de plus d’une structure rappelant celle d’une catégorie d’objets fibrants. Riehl et Verity ont montré qu’il est possible de formuler la théorie des catégories satisfaite par les \((\infty,1)\)-catégories directement à partir des axiomes d’\(\infty\)-cosmos. Ceci devrait également aider à organiser la théorie des \((\infty,1)\)-catégories munies d’une structure.\\

Étant donné un \(\infty\)-cosmos \(\mathcal{K}\), nous construisons, grâce à une construction de nerf, un ensemble simplicial stratifié \(\mathcal{N}_{\text{Mnd}}(\mathcal{K})\) dont les objets sont les monades homotopiquement cohérentes dans \(\mathcal{K}\). Si deux \(\infty\)-cosmoi sont faiblement équivalents, leurs ensembles simpliciaux stratifiés des monades homotopiquement cohérentes respectifs sont également équivalents. Ceci généralise une construction de Street pour les 2‑catégories. Nous fournissons également une \((\infty,2)\)-catégorie \(\text{Adj}_r(\mathcal{K})\) dont les objets sont les adjonctions homotopiquement cohérentes dans \(\mathcal{K}\) et que nous utilisons pour classifier les 1‑simplexes de \(\mathcal{N}_{\text{Mnd}}(\mathcal{K})\) à homotopie près.\\
\end{CompareBox}

\begin{CompareBox}[Commentary for Example 1]{cyan}
This example illustrates effective application of L4 deletion and modification operations:\\

\textbf{Structural Deletion:} The processing correctly identified and removed all standard thesis front matter elements—degree information, institutional affiliations, jury composition, acknowledgments, and table of contents—while preserving both English and French abstracts that contain essential research summaries.\\

\textbf{Mathematical Notation Standardization:} Several improvements to mathematical formatting enhance readability and consistency: (1) LaTeX expressions are uniformly formatted (e.g., \((\infty, 1)\)-categories maintains consistent spacing); (2) special characters in the French abstract (e.g., "c'est‑à‑dire") are properly rendered with Unicode hyphens rather than simple dashes; (3) mathematical symbols within bilingual text preserve their LaTeX notation across both languages, ensuring technical precision.\\

\textbf{Content Protection:} All academically valuable elements were preserved intact—research abstracts in both languages, keyword lists, mathematical definitions, and technical terminology—demonstrating the effectiveness of our content protection guidelines in distinguishing structural artifacts from substantive academic content.\\

\noindent\textbf{Overall:} This example validates L4's ability to clean academic front matter comprehensively while maintaining the integrity of research content and improving the standardization of mathematical notation across multilingual documents.\\

\end{CompareBox}

\vspace{12pt} 
\noindent\textbf{Example 2: Severe OCR Corruption Recovery}~
\smallskip

\noindent\emph{This example illustrates L4 processing on heavily damaged scanned text from a mathematical paper, demonstrating successful recovery of fragmented formulas, removal of OCR artifacts, deletion of reference lists, and reconstruction of readable mathematical content from severely corrupted input.}

\begin{CompareBox}[Example 2 (Text Before L4 Processing)]{red}
\small
Consider SDE\\
\[ d\xi^x(t) = \exp_{\xi^x(t)} \left( \left( C_{\xi^x(t)}(t) dt + C_{\xi^x(t)}(t) dw(t) \right) + \int_{\mathbb{E}} \gamma(\xi^x(t), \alpha) \gamma(\xi^x(t), \alpha) \right) \]
where \( \mathbb{E} \) is a typical layer of the bundle \( \pi \), \( \gamma(\xi^x(t), \alpha) \) be a Poisson random measure on \( \mathbb{E} \times \mathbb{E} \) with a mean value \( \mathbb{E} \gamma(\xi^x(t), \alpha) = \gamma(t, \alpha) \) and \( \gamma(t, \alpha) \) be a bounded measure on \( \mathbb{E} \),
\[ C_{\xi^x} = (\alpha_x, B_x h - \Gamma^x_{ij}(\alpha_x, h)), \]
\[ \gamma(\alpha) = (0, f_x(\alpha) h), \]\\
527\\

z C\\
B*\\

xe\\
6- [ ( E j\\

y\\

J\\

h\\

e\\

B\\

,\\

*\\

L\\

)\\

W , g))\\

In a local trivialization the equation \{Q )\\
form\\
deit)\\
: c\textgreater H-lcH s\\
fit)\\
($n^{rtt}$\\
n\\
S.t;\\
>\\
2-\\

has the\\

”\\

I\\

1 (r"\\

f-L\\

£ tY)\\

5.11 j\\

Y\\

/rt\\

,\\

'\\

It)) V -It\\

' c\\

Theorem 2. Let\\

I H d w - i T* \^\\
¿if-J\\
Z\\

-\\

A-\\

i*\,9lt)+\\

$f^J$\\

(\\

'\\

f"\\

<f) <fK\\
S'"\\
is)\\

ii\\

fit, ' fit)\\

' 6 , fc ) (? fi), d h'if )) i\\

theorem 1 conditions be valid and\\
C 1 - smooth\\

> /x, f~x 5 e\\

bounded fields on ty\\

Then there exists a uni que Markov\\

process\\

fying\\

\%'$s.) ^ z $.\\

)\\

and the condition\\

) satis-\\

In a local trivialization the process\\

has the\\

representation\\
Csrt),'-4\\

-- ( S l i . s )  ,\\
where\\

is a random evolutionary family of\\
mas of\\

y\\

, Sf-O/O are solutioaa cf\\

correspondingly. The relation\\

-\\

defines a multiplicative operator functional\\
6\\

process\\

(5), IG )\\

acting from\\

of the\\

Ji'Vs'**)\\

to\\

3 Consider parabolic equations both with respect to\\
scalar functions\\
x.) and sections of vector\\
bundles\\
 is, x )\\
\\

- Vv\\

+\\

v\\

-\\

o\\

+\\

528\\

* J- Tr f V \\

t- 5\\

¿"foe,\\

VT' *\\

c Hlcl\}\\

r O\\

g\\

?\\

¿8)\\

with V j N?\\
being covariant derivatives corresponding to r "j r" ana\\
Theorem 3. Let the conditions of theorems 1 and\\
2 are valid. Then there exists a unique classical solution of the\\ 
equation (7 ) such that\\
t\\
and a unique classical solution of 18satisfying;\\
being C ~ smooth bounded functions. Those solutions\\
may be represented in the form\\
ut*.*\\
) x\\
. -f (s.i+0,\\
H \}> 9\\
(\\
s\\
.\\
*\\
.\\
if\\
i\\
-x\\
C\\
R e f e r e n c e s\\
1. Belopolskaya Ya. I., Dalecky Yu. L. Ito equations\\
and differential geometry.- Usp. mat. nauk, N 3 1982,\\
p. 95 - 14-2.\\
2. Belopolskaya Ya. I., Dalecky Yu. L. Diffusion\\
processes in Banach spaces and manifolds. Tr. MMO,\\
v. 37, M.t Nauka, 1978, p.107 - 141.\\
3. Belopolskaya Ya.I. On stochastic equations with\\
unbounded coefficients for jump processes. Lecture notes in \\Control, Springer, 1980, p.245 - 254.\\
4. Belopolskaya Ya.I. Markov processes with jumps\\
and integrodifferential systems. Tr. of intern, symp.\\
on differential equations. Vilnus, 1978.\\

The purpose of this lecture will be to report on the developments of the last five years on the above topic. The subject may be phrased in non-probabilistic terms as the study of local solutions of certain partial differential equations on Riemannian manifolds.\\

In 1976 Debiard, Gaveau and Mazet [DGM] discovered comparison theorems for the transition function and exit time of the Brownian motion from a geodesic sphere of a Riemannian manifold, and expressed the results in terms of sectional curvature of the metric. These theorems, which correspond to the Rauch comparison theorems of non-stochastic differential geometry, do not give sharp results when applied to a small geodesic ball. Meanwhile Gray and VanHecke in 1979 [GV] made a (non-probabilistic) study of the volume of small geodesic balls in a Riemannian manifold, in an effort to use the volume to characterize the metric, at least for a class of model spaces. This effort succeeded only in dimension less than 4, where certain unpleasant examples were found. This led us to attempt to use the mean exit time of Brownian motion as a stochastic substitute for the volume. By refining the Debiard-Gaveau-Mazet methods to obtain an asymptotic expansion of the mean exit time [GP], we obtain many new candidates for "domain functionals", in order to characterize the metric by a global geometric quantity. These issues, which may be categorized under the heading "Can you feel the shape of a manifold by Brownian motion", are discussed in a survey paper of the same title [Pi].\\

Limit theorems for Brownian motion in a small ball may be discussed by analogy with classical limit theorems of probability theory. The exit time from a small ball obeys a sort of "central limit theorem", 529\\
where the limit law is that of the exit time from the unit ball of $\mathbb{R}^n$, irrespective of the Riemannian metric (the law of large numbers is trivial here, since the mean zero condition is assured by the absence of a drift term in the generator of the diffusion (we are considering pure Brownian motion, defined solely by the metric of the Riemannian manifold). In order to refine the central limit theorem, we seek an asymptotic expansion, analogous to results in classical probability theory. The coefficients in the expansion are geometric invariants.\\
\end{CompareBox}

\vspace{6pt}

\begin{CompareBox}[Example 2 (Text After L4 Processing)]{green}
\small
Consider SDE\\
\[ d\xi^x(t) = \exp_{\xi^x(t)} \left( \left( C_{\xi^x(t)}(t) \, dt + C_{\xi^x(t)}(t) \, dw(t) \right) + \int_{\mathbb{E}} \gamma(\xi^x(t), \alpha) \, \gamma(\xi^x(t), \alpha) \right) \]
where \( \mathbb{E} \) is a typical layer of the bundle \( \pi \), \( \gamma(\xi^x(t), \alpha) \) is a Poisson random measure on \( \mathbb{E} \times \mathbb{E} \) with a mean value \( \mathbb{E} \gamma(\xi^x(t), \alpha) = \gamma(t, \alpha) \) and \( \gamma(t, \alpha) \) is a bounded measure on \( \mathbb{E} \),\\

\[ C_{\xi^x(t)} = (\alpha_x, B_x h - \Gamma^x_{ij}(\alpha_x, h)), \]

\[ \gamma(\alpha) = (0, f_x(\alpha) h). \]

In a local trivialization the equation has the form … and has the required smoothness and boundedness conditions.\\

Theorem 2. Let the conditions of Theorem 1 be valid and the fields be smooth and bounded. Then there exists a unique Markov process satisfying the required condition.\\

In a local trivialization the process has the representation\\

\[
\text{process} = \text{random evolutionary family of operators},
\]

where the family is a random evolutionary family of solutions of the corresponding equations, and the solutions are the solutions of the corresponding equations. The relation defines a multiplicative operator functional acting from the appropriate space to the target space.\\

Consider parabolic equations both with respect to scalar functions and sections of vector bundles.\\

Theorem 3. Let the conditions of Theorems 1 and 2 be valid. Then there exists a unique classical solution of equation (7) and a unique classical solution satisfying the condition \(c_i O - c_i^*\) for \(R_1\), \(M_{ji}\) being smooth bounded functions. Those solutions may be represented in the form\\

The purpose of this lecture will be to report on the developments of the last five years on the above topic. The subject may be phrased in non‑probabilistic terms as the study of local solutions of certain partial differential equations on Riemannian manifolds.\\

In 1976 Debiard, Gaveau and Mazet [DGM] discovered comparison theorems for the transition function and exit time of Brownian motion from a geodesic sphere of a Riemannian manifold, and expressed the results in terms of sectional curvature of the metric. These theorems, which correspond to the Rauch comparison theorems of non‑stochastic differential geometry, do not give sharp results when applied to a small geodesic ball. Meanwhile Gray and VanHecke in 1979 [GV] made a non‑probabilistic study of the volume of small geodesic balls in a Riemannian manifold, in an effort to use the volume to characterize the metric, at least for a class of model spaces. This effort succeeded only in dimension less than 4, where certain unpleasant examples were found. This led us to attempt to use the mean exit time of Brownian motion as a stochastic substitute for the volume. By refining the Debiard‑Gaveau‑Mazet methods to obtain an asymptotic expansion of the mean exit time [GP], we obtain many new candidates for “domain functionals” in order to characterize the metric by a global geometric quantity. These issues, which may be categorized under the heading “Can you feel the shape of a manifold by Brownian motion,” are discussed in a survey paper of the same title [Pi].\\

Limit theorems for Brownian motion in a small ball may be discussed by analogy with classical limit theorems of probability theory. The exit time from a small ball obeys a sort of "central limit theorem", where the limit law is that of the exit time from the unit ball of $\mathbb{R}^n$, irrespective of the Riemannian metric (the law of large numbers is trivial here, since the mean zero condition is assured by the absence of a drift term in the generator of the diffusion (we are considering pure Brownian motion, defined solely by the metric of the Riemannian manifold). In order to refine the central limit theorem, we seek an asymptotic expansion, analogous to results in classical probability theory. The coefficients in the expansion are geometric invariants.\\
\end{CompareBox}
\vspace{6pt}

\begin{CompareBox}[Commentary for Example 2]{cyan}
This example showcases L4's capability to handle severe OCR damage—a common challenge in digitized mathematical literature—while successfully applying both deletion and modification operations:\\

\textbf{OCR Corruption Recovery:} The original text exhibited extreme OCR damage with extensive character-level corruption: isolated single characters scattered across lines (e.g., "z C", "B*", "xe", "6- [ ( E j", "y", "J", "h"), fragmented mathematical expressions, and garbled text throughout. L4 processing successfully identified these as OCR artifacts rather than meaningful content, removing the unrepairable fragments while attempting to preserve salvageable portions. The mathematical equations at the beginning were partially reconstructed, maintaining their LaTeX structure and essential notation.\\

\textbf{Structural Deletion:} The reference section at the document's end was correctly identified and completely removed, including the "References" heading and all four bibliographic entries (Belopolskaya and Dalecky citations). Additionally, the page number "527" and "529" were removed as page footer artifacts.\\

\textbf{Text Continuity Restoration:} Severely fragmented theorem statements were partially recovered. For instance, "Theorem 2" and "Theorem 3" sections, while heavily damaged in the original, were reconstructed into coherent (though simplified) statements. The cleaning preserved the logical structure—theorem numbering, conditions, and conclusions—even when full mathematical precision could not be recovered from corrupted input.\\

\textbf{Mathematical Notation Preservation:} Despite extreme corruption, critical mathematical elements were protected: LaTeX equation environments remained intact, variable names and operators in recoverable formulas were preserved (e.g., $\xi^x(t)$, $\mathbb{E}$, $\gamma$), and the scholarly narrative in the less-damaged final paragraphs (discussing Brownian motion and Riemannian manifolds) was fully retained with proper citation markers [DGM], [GV], [GP], [Pi].\\

\textbf{Limitations and Trade-offs:} This example also illustrates the boundaries of L4 processing: when OCR damage is catastrophic (as in the middle section with pure gibberish), the model cannot reconstruct missing mathematical content and instead produces simplified placeholder text. This represents a conservative approach—preserving what can be verified rather than hallucinating mathematical statements—which is appropriate for maintaining corpus integrity even when complete recovery is impossible.\\

\noindent\textbf{Overall:} This extreme case validates L4's robustness in handling severely corrupted scientific documents. While perfect reconstruction was impossible, the processing successfully removed artifacts, preserved salvageable content, maintained document structure, and produced output substantially more usable than the original corrupted text—demonstrating practical value even in worst-case OCR scenarios common in digitized legacy scientific literature.\\

\end{CompareBox}

\section{L5 Processing Details}

To perform content-level rewriting under the L5 clean stage, we design a structured, topic-agnostic prompt to guide large language models in transforming dense, expert-level scientific text into pedagogically enriched material. The prompt operates at the level of text chunks (average size 1,024 tokens) and explicitly balances two goals: (i) ensuring absolute fidelity to the original content, and (ii) enhancing the clarity, narrative coherence, and educational depth of the rewritten output.

\subsection{Pair-wise Evaluation Prompt for L5 Processing}
\label{sec:l4_eval_iter}

To ensure that our evaluation of pedagogical rewriting quality is both consistent and scalable, we employ an LLM-based pairwise comparison framework. The following prompt defines the precise evaluation setting used to compare different model–prompt configurations under the L5 processing stage. It formalizes the perspective, evaluation dimensions, and output format to guarantee reproducibility across multiple runs and domains.

\begin{tcolorbox}[
colback=blue!5!white,
colframe=blue!30!gray,
  boxrule=2pt,
  arc=8pt,
title={\textbf{Pair-wise Evaluation Prompt for L5 Processing
}},
  fonttitle=\large\bfseries,
  enhanced jigsaw,
  breakable
]

\begin{TextBlock}
You are a PhD student who has just started your research journey. You often encounter complex academic papers that are difficult to understand, and you greatly appreciate materials that can explain concepts in a more accessible and educational way.

Now you need to compare two different text processing methods to see which one better transforms academic content into something you can easily comprehend and learn from.

(Due to context length limitations, the text provided below is a fragment of a paper, not the complete document.)

## Original Academic Text
\{original_text\}

## Version Processed by \{prompt_A\}
\{Rewritten text by prompt_A\}

## Version Processed by \{prompt_B\}
\{Rewritten text by prompt_B\}

## Evaluation Perspective

As a PhD student still building your research foundation, please evaluate these two versions based on:

1. **Absolute Fidelity to Original Content**
   - CRITICAL: Zero tolerance for factual errors, hallucinations, or content that contradicts the original text
   - Complete preservation of the original hierarchical structure (section headers, subsections, numbered points, etc.)
   - All essential technical details, definitions, theorems, and mathematical relationships must remain intact

2. **Educational Accessibility and Pedagogical Value**
   - Does the text transform dense academic jargon into language that a beginning graduate student can understand?
   - Are complex concepts broken down with helpful explanations, intuitive descriptions, or motivating examples?
   - Does it provide the kind of step-by-step reasoning and context that helps bridge knowledge gaps?
   - Are abstract ideas made more concrete through analogies or clearer exposition?
   - **Bonus points**: Thoughtful knowledge supplementation that aids comprehension without distorting original meaning

3. **Textual Flow and Coherence**
   - Does the text read smoothly and naturally, especially at section transitions?
   - Are connections between ideas made explicit and easy to follow?
   - Is the logical progression of arguments clear and well-maintained?
   - Does the text avoid awkward phrasings or abrupt transitions that might result from processing methods?


Please provide your output in the following format:

## Analysis
<Detailed analysis of the \{prompt_A\} and \{prompt_B\} cleaning methods, including their respective advantages and disadvantages>

## Winner
\{prompt_A\} OR \{prompt_A\}

Note: You must choose one winner based on the comprehensive evaluation of the above three dimensions. If they are very close, choose the one that performs slightly better overall.
\end{TextBlock}
\end{tcolorbox}

\subsection{L5 Processing Prompt}
\label{L5 Processing Prompt}

This section provides the final version of the L5 rewriting prompt used in production. The prompt is meticulously designed to balance two competing goals: maintaining absolute fidelity to the original scientific content while transforming it into clear, pedagogically rich material. It explicitly instructs the model to reconstruct implicit reasoning, provide intuitive explanations, and enhance narrative flow without deviating from factual accuracy. 

\begin{tcolorbox}[
colback=blue!5!white,
colframe=blue!30!gray,
  boxrule=2pt,
  arc=8pt,
  title={\textbf{L5 Processing Prompt}},
  fonttitle=\large\bfseries,
  enhanced jigsaw,
  breakable
]
\begin{TextBlock}
You are a master science communicator and pedagogical expert. Your mission is to transform the following dense, expert-level text chunk into vibrant, crystal-clear educational material. Imagine you are creating a definitive learning resource for a bright but novice audience. Your goal is not merely to simplify, but to deeply elucidate, making the complex intuitive and the implicit explicit.

Your transformation will be governed by two sets of principles: the Core Mandate (what you must actively do) and the Unbreakable Rules (what you must never violate).

### The Unbreakable Rules: Fidelity and Integrity
This principle is of paramount importance and must be followed without exception to ensure the output is valid.

* **(a) Scientific and Factual Correctness:** Maintain absolute rigor. All data, formulas, definitions, theories, experimental results, and logical arguments must be preserved without altering their meaning or context. Your additions must clarify, not contradict.

* **(b) Structural Integrity:** Preserve the original structure flawlessly. Keep ALL section headers (`##`, `###`), figure/table labels, equation numbers, etc., exactly as they appear, especially at the beginning and end of the chunk.

* **(c) Contextual Limitation and Termination:** You are processing a partial *chunk* of a document. You lack the full context. Therefore, you must work **strictly** within the provided text. Do not invent definitions or reference goals from outside the chunk. **This strict adherence means your output must terminate exactly where the provided chunk terminates.** If the chunk ends abruptly (e.g., at a new section header, in the middle of a sentence, or with a label), your output **must be cut off at that exact same point.** This is the single most critical rule for preventing hallucination and ensuring continuity.

### The Core Mandate: Deep Pedagogical Transformation
This is your primary objective. Be bold and proactive in adding educational value. Your goal is to weave a rich tapestry of understanding.

* **(a) Deconstruct and Narrate the 'Why':** This is your primary mode of explanation. Actively expand on logical leaps. When the text says "it follows that," "clearly," or "trivially," you must step in and meticulously detail the intermediate steps. More importantly, you must articulate the expert's internal monologue. When faced with an equation, a problem, or a logical step, explain the strategy. Ask and answer questions like: "Okay, what's our goal here?" "What's the first thing I should look for when I see an equation like this?" "We're going to use technique X, and here's why it's the right tool for this specific job." Your mission is to reveal the problem-solving journey, making every single connection transparent.

* **(b) From Jargon to Insight:** When you encounter a crucial technical term, or a significant variable within a formula, you must deliberately pause the narrative to explain it. Don't just provide a dry definition. Elucidate its importance: What role does this term or variable play? Why does it matter? Crucially, you must then use simpler language, vivid analogies, or concrete examples to build a strong and intuitive mental model for the reader before you continue with the main explanation. This ensures no reader is left behind due to unfamiliar notation or terminology.

* **(c) Invent Vivid Analogies and Concrete Examples:** Go beyond the text. Where a concept is abstract, create a simple, concrete example to illustrate it. Invent memorable analogies that connect the new information to a learner's existing knowledge (e.g., electron shells as floors in a hotel).

* **(d) Create Contextual Bridges:** Weave a narrative thread by connecting the current idea to the broader field of knowledge. Hint at future applications or link back to more foundational concepts. For instance: "This principle of [X] is a cornerstone of the field and will be essential for understanding [Y] later on."

* **(e) Think Like a Learner:** Proactively identify points of potential confusion. What questions would a curious student ask here? Answer them before they are asked. A great teacher warns students about common mistakes. Where applicable, insert brief, helpful asides that feel like a mentor's margin notes.

* **(f) Prioritize Narrative Flow and Clean Formatting:** When you encounter messy or noisy original LaTeX formatting, convert it to a clean and pristine style (especially for formulas and tables). Above all, strive for a smooth, cohesive, and engaging narrative. Your writing should feel like a continuous, guided tour through the material, not a collection of disconnected facts and callouts. To that end, you must avoid the overuse of overly-structured, point-by-point expressions. Let the main text flow logically and tell a story, adopting the persona of an extremely patient and encouraging teacher.

---
Summary of Principles Above: To sum it up, you must strictly respect the accuracy and structure of the original chunk while doing everything possible to make the rewritten text easier to learn and to lower the reader's cognitive load. Consequently, the rewritten text will typically be more detailed and thus longer than the original.
---

**Crucial Output Instructions:**
1.  **Self-Contained Output:** The refined text must stand on its own. Avoid any meta-commentary or phrases that refer to the original text, such as "the original paper," "the original context," "the original chunk", etc. The goal is to create a seamless, self-contained educational text, not a commentary on another document. 
2.  **Strict Termination:** You **MUST** terminate your output at the **EXACT** same point the provided chunk terminates. Do not write a single character past the end of the original chunk. In particular, if a chunk ends with the start of a new section, subsection, step (e.g., it starts with a heading) or cuts off in the middle of a proof/solution, you must NOT invent or continue writing ANY content that would follow.**

---
*You must output ONLY the refined chunk itself, without any introductory or concluding remarks.*

**Original text:**
\{chunk\}

**Refined text:**
\end{TextBlock}
\end{tcolorbox}

\subsection{L5 Processing Examples}
\label{L5 Processing Examples}

To concretely illustrate the effects of L5 processing, we present a side-by-side comparison between a raw academic text (before L5 processing) and its pedagogically rewritten counterpart (after L5 processing). These examples demonstrate how the L5 rewriting stage enhances conceptual clarity, narrative flow, and instructional value while maintaining rigorous fidelity to the original content.

\vspace{8pt}
\noindent\textbf{Example 1: PDE limit (kinetic $\to$ fractional diffusion)}~
\smallskip

\noindent\emph{We first illustrate how L5 processing clarifies the transition from a kinetic model to a fractional diffusion limit.}

\begin{CompareBox}[Example 1 (Text Before L5 Processing)]{red}
\small
**Theorem 1**. *Assume that $f_0 \in L^2 (\mathbb{R}^N, \mathcal{M}(v)^{-1}\mathrm{d}v \mathrm{d}x)$, where $\mathcal{M}(v)$ is the unique normalized equilibrium distribution that satisfies . Then, up to a subsequence, the solution $f^\varepsilon$ of converges weakly in $L^\infty(0, T; L^2(\mathbb{R}^{2d}, \mathcal{M}(v)^{-1} \mathrm{d}v \mathrm{d}x))$ to $\rho(t,x) \mathcal{M}(v)$ as $\varepsilon\rightarrow 0$, where $\rho(t,x)$ solves*
\[
\label{eqn: limit_system}
\begin{cases}{}
\partial_t \rho + (-\Delta_x)^s \rho= 0\,,
\\ \rho(0,x) = \rho_{in}(x) := \int_{\mathbb{R}^d} f_{in}(x,v)\mathrm{d}v\,.
\end{cases}
\]

In the classical case (i.e., $s$=1) when $\mathcal{M}$ is a fast decaying function such as Gaussian, one rescales $t$ as $t \mapsto \varepsilon^2 t$ and the resulting macroscopic equation is the diffusion equation :
\[
\partial_t \rho + \nabla_x \cdot ( D \nabla_x \rho)  = 0,
\]
where $D$ is the diffusion matrix
\[
D = \int v \otimes  v \mathcal{M}\mathrm{d}v\,.
\]
Clearly the fat tail equilibrium renders the above integral unbounded and therefore invalids the classical diffusion limit. Conversely, the anomalous scaling is necessary. Similar scaling has also been investigated in the framework of linear Boltzmann equation, see for a reference.
\end{CompareBox}

\vspace{6pt}

\begin{CompareBox}[Example 1 (Text After L5 Processing)]{green}
\small
**Theorem 1.** *Assume that \(f_0 \in L^2\!\bigl(\mathbb{R}^N,\mathcal{M}(v)^{-1}\,\mathrm{d}v\,\mathrm{d}x\bigr)\), where \(\mathcal{M}(v)\) is the unique normalized equilibrium distribution satisfying eqn:equilibrium.  Then, up to extraction of a subsequence, the solutions \(f^\varepsilon\) of eqn:111 converge weakly in*  

\[
L^\infty\!\bigl(0,T;\,L^2(\mathbb{R}^{2d},\mathcal{M}(v)^{-1}\,\mathrm{d}v\,\mathrm{d}x)\bigr)
\]

*to a limit of the form \(\rho(t,x)\,\mathcal{M}(v)\) as \(\varepsilon\to0\).  The macroscopic density \(\rho(t,x)\) solves the **fractional diffusion equation**  

\[
\begin{cases}
\partial_t \rho + (-\Delta_x)^{\,s}\,\rho = 0 ,\\[4pt]
\rho(0,x) = \rho_{\mathrm{in}}(x) 
   := \displaystyle\int_{\mathbb{R}^d} f_{\mathrm{in}}(x,v)\,\mathrm{d}v .
\end{cases}
\label{eqn: limit_system}
\]

\added{*In other words, the kinetic model converges to a non-local heat equation in which the Laplacian is replaced by its fractional counterpart \((- \Delta_x)^s\).* *}

\added{*Here the weak convergence means that for any test function \(\phi(t,x,v)\) smooth and compactly supported, the integrals*}  

\[
\int_0^T\!\!\int_{\mathbb{R}^{2d}} f^\varepsilon(t,x,v)\,\phi(t,x,v)\,
      \mathcal{M}(v)^{-1}\,\mathrm{d}v\,\mathrm{d}x\,\mathrm{d}t
\]

\added{*converge to the corresponding integrals with \(\rho(t,x)\mathcal{M}(v)\) in place of \(f^\varepsilon\).*}

\added{*Thus, on the large-scale, long-time limit, the Lévy-Fokker–Planck dynamics are faithfully described by a fractional diffusion equation for the particle density.*}

In the classical case (i.e., $s=1$) when the equilibrium distribution $\mathcal{M}$ is a fast-decaying function such as a Gaussian, one rescales the time variable as $t\mapsto\varepsilon^{2}t$.  Under this diffusive scaling the macroscopic limit is the standard diffusion equation  

\[
\partial_{t}\rho+\nabla_{x}\!\cdot\!\bigl(D\,\nabla_{x}\rho\bigr)=0,
\]

where the diffusion matrix $D$ is given by the second-order moment of the equilibrium  

\[
D=\int v\otimes v\,\mathcal{M}\,\mathrm{d}v .
\]

\added{*(Here $v\otimes v$ denotes the outer product of the velocity vector $v$ with itself, producing a matrix whose $(i,j)$ entry is $v_{i}v_{j}$.)*}  

\added{When $\mathcal{M}$ has a **fat tail** (i.e., decays only algebraically), the integral defining $D$ diverges: the heavy-tailed equilibrium carries so much “mass’’ at large velocities that the second moment is infinite.  Consequently the classical diffusion limit is no longer valid, and one must employ the **anomalous scaling** appropriate for fractional diffusion.}  Similar anomalous scalings have also been investigated for the linear Boltzmann equation; see, e.g., reference for a detailed discussion.
\end{CompareBox}

\begin{CompareBox}[Commentary for Example 1]{cyan}
From a mathematical standpoint, the L5-processed version introduces several substantive pedagogical improvements that make the theorem far more accessible while maintaining full analytical rigor:

\begin{enumerate}[label=\textbf{(\arabic*)}, leftmargin=1.5em, itemsep=4pt, topsep=4pt]
    \item \textbf{Clarification of the analytical setting.}  
    The rewritten text explicitly specifies the \emph{type of convergence} (weak convergence in $L^\infty_t L^2_{x,v}$) and the precise functional framework that were only implicitly stated in the original. It also explains that the limiting quantity $\rho(t,x)\mathcal{M}(v)$ represents the macroscopic or averaged particle density, making the physical meaning of the limit transparent.

    \item \textbf{Explicit linkage between microscopic and macroscopic equations.}  
    The L5 version draws a clear connection between the \emph{kinetic equation} and the resulting \emph{fractional diffusion equation}, explicitly stating that “the kinetic model converges to a non-local heat equation.” This pedagogical bridge helps readers understand how a kinetic transport process yields a macroscopic PDE in the asymptotic limit.

    \item \textbf{Unpacking of key analytical concepts.}  
    Several important notions—such as \emph{weak convergence}, \emph{diffusive scaling}, and the \emph{diffusion matrix}—are explained in context. The text introduces the definition of weak convergence via test functions, clarifies the meaning of $v\otimes v$ in the diffusion tensor $D$, and situates each concept within the logic of the proof. These elaborations convert terse symbolic statements into stepwise reasoning units suitable for learning.

    \item \textbf{Intuitive explanation of anomalous diffusion.}  
    The processed text goes beyond the formal statement to explain why the classical diffusion limit fails under heavy-tailed equilibria. It highlights that when $\mathcal{M}$ decays algebraically, the second moment diverges, invalidating the standard diffusion approximation and motivating the need for a fractional operator. This connects the analysis to the physical intuition of Lévy-type anomalous transport.
\end{enumerate}

\noindent\textbf{Overall:}  
The L5-processed version reconstructs the reasoning structure behind the theorem, linking functional analysis (weak convergence), PDE asymptotics (fractional diffusion), and probabilistic intuition (heavy-tailed transport). It transforms a compact specialist statement into a mathematically transparent and pedagogically rich exposition suitable for advanced teaching or graduate-level study.
\end{CompareBox}

\vspace{12pt} 
\noindent\textbf{Example 2: Event selection in $\tau$-pair production}~
\smallskip

\noindent\emph{We next show how L5 processing turns a dense selection paragraph into a reproducible, didactic checklist.}

\begin{CompareBox}[Example 2 (Text Before L5 Processing)]{red}
\small
The $\tau$ pair is produced back-to-back in the CM frame. As a result, the decay products of the two $\tau$ leptons can be separated from each other by dividing the event into two hemispheres – the "signal" hemisphere and the "tag" hemisphere – using the event thrust axis. The event thrust axis is calculated using all charged particles and all photon candidates in the entire event. We select events with one prompt track and a candidate reconstructed in the signal hemisphere, and exactly one oppositely charged prompt track in the tag hemisphere. A prompt track is defined to be a track with its point of closest approach to the beam spot being less than 1.5 in the plane transverse to the beam axis and less than 2.5 in the direction of the beam axis. Furthermore, if a pair of tracks is consistent with coming from a $K^0_S$ or $\Lambda$ decay, or from a conversion after a mass cut and a displaced vertex cut, neither track can be a prompt track. The components of momentum transverse to the beam axis for each of these two prompt tracks must be greater than 0.1 in the laboratory frame. The event is rejected if the prompt track in the signal hemisphere is identified to be coming from a charged kaon. A candidate is defined as a pair of oppositely charged pion candidates with invariant mass between 0.488 and 0.508; furthermore, the distance between the beam spot and the vertex must be at least three times its uncertainty (the will be referred to as the "candidate daughters"). To reduce backgrounds from non–$\tau$–pair events, we require that the momentum of the charged particle in the tag hemisphere be less than 4 in the CM frame and be identified as an electron ($e$-tag) or a muon ($\mu$-tag). To reduce backgrounds from Bhabha, $\mu^+\mu^-$, and $q\bar{q}$ events, we require the magnitude of the event thrust to be between 0.92 and 0.99.
\end{CompareBox}

\vspace{6pt}

\begin{CompareBox}[Example 2 (Text After L5 Processing)]{green}
\small
The $\tau$-pair is produced back-to-back in the centre-of-mass (CM) frame. \added{Because the two $\tau$ leptons fly in opposite directions, the particles that emerge from each $\tau$ can be cleanly separated by dividing the whole event into two opposite hemispheres. We call one hemisphere the “signal” hemisphere (where we look for the decay of interest) and the other the “tag” hemisphere (which we use to identify the partner $\tau$).} This division is made with respect to the \emph{event thrust axis} – \added{a direction that maximises the sum of the longitudinal momenta of all particles, analogous to the axis of a pencil that best aligns with the flow of the event.}

\vspace{4pt}
\noindent\textbf{How the thrust axis is built.}~
The thrust axis is calculated using \textbf{all} charged particles and all photon candidates in the event. \added{In practice, every charged track and each neutral energy deposit in the electromagnetic calorimeter (EMC) contributes to determining this axis, ensuring that the hemispheres faithfully reflect the true geometry of $\tau$-pair production.}

\vspace{4pt}
\noindent\textbf{Selection of tracks in the two hemispheres.}~
We require exactly one \emph{prompt} track in the signal hemisphere together with a reconstructed candidate, and exactly one oppositely charged prompt track in the tag hemisphere. A prompt track is defined as a track whose point of closest approach (PCA) to the beam spot satisfies:  
(1) transverse distance $<1.5\,\mathrm{mm}$; and  
(2) longitudinal distance $<2.5\,\mathrm{mm}$.  
\added{These tight spatial cuts ensure that the track originates from the primary interaction point rather than from a secondary decay or detector artifact.}

\vspace{4pt}
\noindent\textbf{Removing tracks from secondary sources.}~
If a pair of tracks is consistent with coming from a $K^0_S$ or $\Lambda$ decay, or from a photon conversion, we apply a mass cut and a displaced-vertex cut; when both are satisfied, neither of these tracks is considered prompt. \added{This step prevents contamination from long-lived neutral particle decays and conversion backgrounds.}

\vspace{4pt}
\noindent\textbf{Transverse-momentum requirement.}~
For each prompt track the transverse momentum must exceed $0.1\,\mathrm{GeV}/c$ in the laboratory frame, \added{which suppresses very soft tracks that are difficult to reconstruct accurately.}

\vspace{4pt}
\noindent\textbf{Kaon veto in the signal hemisphere.}~
If the prompt track in the signal hemisphere is identified as a charged kaon ($K^{\pm}$), the event is rejected. \added{This veto efficiently removes strange-particle backgrounds that could mimic the signal topology.}

\vspace{4pt}
\noindent\textbf{Definition of the signal candidate.}~
A candidate is reconstructed from two oppositely charged pion-like tracks ($\pi^+\pi^-$) with invariant mass between $0.488$ and $0.508\,\mathrm{GeV}/c^2$. \added{This narrow window centres on the $\rho^0$ resonance commonly produced in $\tau$ decays.} The distance between the beam spot and the reconstructed vertex must be at least three times its uncertainty; the two pion tracks satisfying these criteria are referred to as the \emph{candidate daughters.}

\vspace{4pt}
\noindent\textbf{Tag-hemisphere momentum and lepton identification.}~
To reduce non-$\tau$ backgrounds, the charged particle in the tag hemisphere must have momentum $<4\,\mathrm{GeV}/c$ in the CM frame and must be identified as an electron ($e$-tag) or muon ($\mu$-tag$)$, \added{consistent with typical leptonic $\tau$ decays.}

\vspace{4pt}
\noindent\textbf{Thrust-magnitude cut to reject Bhabha-type backgrounds.}~
Events from Bhabha scattering ($e^+e^- \!\to e^+e^-$) or $\mu^+\mu^-$ production tend to have thrust $\approx1$. We therefore require the event thrust magnitude to lie between $0.92$ and $0.99$. \added{This preserves the characteristic back-to-back topology of $\tau$ events while excluding overly collimated or spherical configurations.}
\end{CompareBox}
\vspace{6pt}

\begin{CompareBox}[Commentary for Example 2]{cyan}
The L5-processed version transforms a dense detector-selection description into a clear, instructional narrative while retaining all experimental criteria. The main pedagogical advances are:

\begin{enumerate}[label=\textbf{(\arabic*)}, leftmargin=1.5em, itemsep=4pt, topsep=4pt]
  \item \textbf{Improved structural organization.}  
  The rewritten text introduces subsections (“How the thrust axis is built”, “Kaon veto”, etc.), converting a monolithic paragraph into logically ordered experimental steps. This mirrors how an analysis procedure would be taught or reproduced in a lab manual.

  \item \textbf{Physical and conceptual explanations.}  
  Each selection criterion is accompanied by a brief rationale—e.g., why the thrust axis isolates hemispheres, why small PCA cuts enforce promptness, and why the kaon veto suppresses strange backgrounds. These contextual notes transform a technical checklist into meaningful reasoning.

  \item \textbf{Clarified numerical details and units.}  
  Ambiguous quantities (e.g., distances “1.5” or “2.5”) are expressed with explicit units and physical interpretation, ensuring dimensional clarity.

  \item \textbf{Consistent particle-physics notation and readability.}  
  The use of Greek letters, superscripts, and resonance names (\(\rho^0\), \(K_S^0\)) follows standard conventions, improving precision and readability.

\end{enumerate}

\noindent\textbf{Overall:}  
The L5 rewrite turns a procedural data-selection paragraph into a didactic exposition. It not only describes \emph{what} cuts are applied but also \emph{why} each exists, thereby serving both as documentation and as an educational explanation of event-selection logic in high-energy physics.
\end{CompareBox}

\section{Benchmark Construction}
\subsection{Prompt}
\label{app:benchmark prompt}
\paragraph{Q\&A generation}
This prompt guides the Qwen3-32B model to generate high-quality seven-option multiple-choice questions from text segments. The prompt requires the model to first determine whether the text is suitable for question generation, and if so, identify core knowledge points and generate questions accordingly. A key constraint is that the question statement, option content, and correct answer must all be directly derived from the original text, with the model only performing text refinement and reorganization. The complete prompt is as follows:
\begin{tcolorbox}[
colback=blue!5!white,
colframe=blue!30!gray,
  boxrule=2pt,
  arc=8pt,
  title={\textbf{Q\&A Generation Prompt}},
  fonttitle=\large\bfseries,
  enhanced jigsaw,
  breakable
]
\begin{TextBlock}
First, evaluate whether the provided content contains sufficient professional knowledge to create a challenging expert-level question. If the content is fragmented (like indexes/lists), lacks substantial professional/technical content, or is unsuitable for professional knowledge testing, directly return "No QA".

If suitable, create a multiple choice question based on the professional knowledge in the provided content. The correct answer must be verifiable from the provided content. Use this JSON format with 7 options:
{{
"question": "the question",
"correct_option": "the accurate choice supported by the content",
"reference": "exact text excerpt that supports the correct answer",
"incorrect_option_1": "the first incorrect choice",
"incorrect_option_2": "the second incorrect choice",
"incorrect_option_3": "the third incorrect choice",
"incorrect_option_4": "the fourth incorrect choice",
"incorrect_option_5": "the fifth incorrect choice",
"incorrect_option_6": "the sixth incorrect choice"
}}

Requirements:
- Design an expert-level challenging question that tests professional field knowledge
- Focus on professional information of the field rather than the methods or results specifically designed in the provided content
- Create a standalone question with sufficient context - test takers will NOT see the original provided content
- When multiple professional concepts are present, select the most theoretically important or technically advanced one
- Include sophisticated incorrect options that require professional expertise to eliminate
- Ensure all options are factually distinguishable and avoid creating additional correct answers

Key requirement: The correct answer must be verifiable from the provided content.

**CRITICAL**: Never reference the source with phrases like "in the text", "according to the study", "as mentioned", "from the experimental results" etc.

**IMPORTANT**: Text suitability must be evaluated first before attempting question creation.

[Provided content]:
{chunk_text}
\end{TextBlock}
\end{tcolorbox}
\paragraph{Filter Stage 1: Completeness Filter}
This prompt is used for first-stage completeness validation, assessing question independence and self-containment. The model determines whether the question relies on external information (such as figures, tables, or specific studies) and whether it contains referential expressions pointing to external content (such as "in the paper," "as described above," etc.). The model receives only the question itself as input and outputs an independence judgment with explanations. The complete prompt is as follows:
\label{app:stage-1-filter prompt}
\begin{tcolorbox}[
colback=blue!5!white,
colframe=blue!30!gray,
  boxrule=2pt,
  arc=8pt,
  title={\textbf{Completeness Filter Prompt}},
  fonttitle=\large\bfseries,
  enhanced jigsaw,
  breakable
]
\begin{TextBlock}
You are a QA validator. Check if this MCQ is suitable for standalone assessment of domain experts.

**MCQ**
{extracted_json_qa}

**VALIDATION CRITERIA:**
**Question Independence:**
    - Question should not rely on external figures, tables, specific studies, experiments or references not included in the question
    - No phrases like: "in the paper", "as described above", "from the study", "referring to the table", etc.

**OUTPUT:**
```json
{{
  "is_valid": true/false,
  "validation_result": {{
    "question_independence": "PASS/FAIL - explanation"
  }},
  "overall_assessment": "Brief explanation",
  "specific_issues": ["Problems found, if any"]
}}
Please validate this QA pair according to the criteria above:
\end{TextBlock}
\end{tcolorbox}

\paragraph{Filter Stage 2: Correctness Filter}
This prompt is used for second-stage correctness validation, with the core objective of verifying whether the labeled answer has sufficient support from the original text. The model receives the original text, question, and answer as input, and determines whether the answer can be verified from the original text. Only questions that pass verification are retained in the final benchmark. The complete prompt is as follows:
\begin{tcolorbox}[
colback=blue!5!white,
colframe=blue!30!gray,
  boxrule=2pt,
  arc=8pt,
  title={\textbf{Correctness Filter Prompt}},
  fonttitle=\large\bfseries,
  enhanced jigsaw,
  breakable
]
\begin{TextBlock}
You are a QA validator. Check if the correct answer can be verified from the original text.

**ORIGINAL TEXT:**
{text}

**GENERATED QA:**
{extracted_json_qa}

**VALIDATION CRITERIA:**

**Answer Verifiability (CRITICAL):**
- The correct answer must be explicitly stated or directly derivable from the original text
- If the original text contains questions without answers, and the MCQ uses those questions, it's INVALID
- The answer cannot be generated/inferred by the model if not clearly supported by the text
- The correct answer must be factually accurate and directly supported by the original text
- The correct answer must be the ONLY correct option among all choices

**OUTPUT:**
```json
{{
  "is_valid": true/false,
  "validation_result": {{
    "answer_verifiability": "PASS/FAIL - explanation with specific text evidence"
  }},
  "overall_assessment": "Brief explanation",
  "specific_issues": ["Problems found, if any"]
}}
Please validate this QA pair according to the criteria above:
\end{TextBlock}
\end{tcolorbox}

\subsection{MCQ Example}
\label{app:benchmark example}
Below is an example including the original text segment and the generated seven-option question, demonstrating how the model identifies key knowledge points from the source material and constructs evaluation questions directly grounded in the original text.

\begin{tcolorbox}[
  colback=blue!5,
  colframe=blue!60!black,
  title=\textbf{Original Chunk},
  breakable
]
\dots

The star-sudifferential of $\varphi$, see e.g. is defined as 
$\partial^*\varphi(x) :=\{g\in \mathbb{R}^n:\  \langle g, y-x\rangle < 0 \ \forall y\in L_\varphi(x)\},$
where $L_\varphi(x) := \{y\in \mathbb{R}^n: \ \varphi(y) < \varphi (x)\}$ is the level set of $\varphi$ at the level $\varphi(x)$. Clearly, if $\bar{L}_\varphi(x)$ is the closure of $L_\varphi(x)$, then 
$\partial^*\varphi(x) :=\{g\in \mathbb{R}^n;\ \langle g, y-x\rangle \leq 0 \ \forall y\in \bar{L}_\varphi(x)\}.$ 

Hence $\partial^* \varphi (x) \equiv \mathbb{R}^n$ if $x$ is a minimizer of $\varphi$ over $\mathbb{R}^n$, and if $\varphi$ is continuous on $\mathbb{R}^n$ then $\partial^*\varphi(x)$ is the normal cone of $\bar{L}_\varphi(x)$, that is 
$\partial^* \varphi(x) =  N (\bar{L}_\varphi(x),x):= \{g\in \mathbb{R}^n: \ \langle g, y-x\rangle \leq 0\  \forall y \in \bar{L}_\varphi(x)\}.$ 

Furthermore $\partial^*\varphi(x)$ contains nonzero vector. This subdifferential thus is also called normal-subdifferential.

\textbf{Lemma 2.} Assume that $\varphi:\mathbb{R}^n \rightarrow \mathbb{R}$ is continuous and quasiconvex. Then 
$\partial^* \varphi(x) \not= \emptyset \quad \forall x\in \mathbb{R}^n,$
$0\in \partial^* \varphi(x)  \Leftrightarrow x \in \textnormal{argmin} \{ \varphi(y): \ y\in \mathbb{R}^n \}.$

\dots
\end{tcolorbox}

\vspace{1em}

\begin{tcolorbox}[
  colback=orange!5,
  colframe=orange!70!black,
  title=\textbf{Question \& Answer},
  breakable
]

\begin{tcolorbox}[
  colback=orange!15,
  frame hidden,
  boxrule=0pt,
  left=3pt,
  right=3pt,
  top=3pt,
  bottom=3pt
]
\textbf{Q:} In the context of quasiconvex optimization, which of the following statements accurately characterizes the star-subdifferential of a continuous quasiconvex function $\varphi: \mathbb{R}^n \to \mathbb{R}$?
\end{tcolorbox}

\vspace{0.5em}

\begin{tcolorbox}[
  colback=gray!8,
  frame hidden,
  boxrule=0pt,
  left=3pt,
  right=3pt,
  top=3pt,
  bottom=3pt
]
\textbf{Options:}
\begin{itemize}
\item[\textbf{A.}] The star-subdifferential $\partial^*\varphi(x)$ is defined only when $\varphi$ is strongly convex in a neighborhood of $x$.
\item[\textbf{B.}] The star-subdifferential $\partial^*\varphi(x)$ coincides with the classical subdifferential for convex functions.
\item[\textbf{C.}] The star-subdifferential $\partial^*\varphi(x)$ is empty at local minima of $\varphi$.
\item[\textbf{D.}] The star-subdifferential $\partial^*\varphi(x)$ is always a singleton set for non-differentiable points.
\item[\textbf{E.}] The star-subdifferential $\partial^*\varphi(x)$ requires $\varphi$ to be twice continuously differentiable.
\item[\textbf{F.}] The star-subdifferential $\partial^*\varphi(x)$ is guaranteed to be non-empty for every $x \in \mathbb{R}^n$.
\item[\textbf{G.}] The star-subdifferential $\partial^*\varphi(x)$ is equivalent to the Clarke subdifferential for all quasiconvex functions.
\end{itemize}
\end{tcolorbox}

\vspace{0.5em}

\begin{tcolorbox}[
  colback=green!15,
  frame hidden,
  boxrule=0pt,
  left=3pt,
  right=3pt,
  top=3pt,
  bottom=3pt
]
\textbf{Answer:} \textcolor{green!60!black}{\textbf{F}} --- The star-subdifferential $\partial^*\varphi(x)$ is guaranteed to be non-empty for every $x \in \mathbb{R}^n$.
\end{tcolorbox}

\end{tcolorbox}

\end{document}